\pdfoutput=1

\documentclass[11pt]{article}
\usepackage{adjustbox}
\usepackage{multirow} 
\usepackage[preprint]{acl}
\usepackage{makecell}
\usepackage{times}
\usepackage{latexsym}
\usepackage{multirow}

\usepackage[T1]{fontenc}

\usepackage[utf8]{inputenc}

\usepackage{microtype}

\usepackage{inconsolata}

\usepackage{graphicx}
\usepackage{graphicx}
\usepackage{booktabs}
\usepackage{array}
\usepackage{amssymb}
\usepackage{amsmath} 
\usepackage{hyperref}
\usepackage{multirow} 
\usepackage{makecell}
\usepackage{xcolor}
\usepackage{colortbl}
\usepackage{xspace}

\usepackage{booktabs}

\definecolor{LightCyan}{rgb}{0.88,1,1}

\newcommand*{\yoruba}{Yor\`ub\'a\xspace}

\title{SSA-COMET: Are LLMs Better at Evaluating Machine Translation Outputs for Under-resourced Sub-Saharan African Languages?}

\title{SSA-COMET: Do LLMs Outperform Learned Metrics in Evaluating MT for Under-Resourced African Languages?}

\author{
    Senyu Li\thanks{$^{}$Equal contribution.}$^{1,2}$\qquad 
    Jiayi Wang\footnotemark[1]$^3$\qquad Felermino D. M. A. Ali$^{7,8}$\qquad \textbf{Colin Cherry}$^5$
    \\  \textbf{Daniel Deutsch}$^5$\qquad \textbf{Eleftheria Briakou}$^5$ \ \ \  \ \ \ \ \textbf{Rui Sousa-Silva}$^{8}$\qquad \\
    \textbf{Henrique Lopes Cardoso}$^{7}$\qquad
    \textbf{Pontus Stenetorp}$^{3,6}$\qquad 
    \textbf{David Ifeoluwa Adelani}$^{1,2,4}$\\
    $^1$Mila - Quebec AI Institute, $^2$McGill University, $^3$University College London,\\  $^4$Canada CIFAR AI Chair, $^5$Google, $^6$LLMC, National Institute of Informatics, \\
    $^7$LIACC, Faculdade de Engenharia, Universidade do Porto, \\
    $^8$CLUP, Faculdade de Letras, Universidade do Porto \\
    \texttt{\{senyu.li, david.adelani\}@mila.quebec} \\\texttt{jiaywang@cs.ucl.ac.uk}\qquad
    \texttt{up202100778@fe.up.pt}
    \\
    }

\begin{document}
\maketitle
\begin{abstract}
Evaluating machine translation (MT) quality for under-resourced African languages remains a significant challenge, as existing metrics often suffer from limited language coverage and poor performance in low-resource settings.
While recent efforts, such as AfriCOMET, have addressed some of the issues, they are still constrained by small evaluation sets, a lack of publicly available training data tailored to African languages, and inconsistent performance in extremely low-resource scenarios.
In this work, we introduce \textit{\textbf{SSA-MTE}}, a large-scale human-annotated MT evaluation (MTE) dataset covering 14 African language pairs from the News domain, with over 73,000 sentence-level annotations from a diverse set of MT systems. Based on this data, we develop \textit{\textbf{SSA-COMET}} and \textit{\textbf{SSA-COMET-QE}}, improved reference-based and reference-free evaluation metrics. We also benchmark prompting-based approaches using state-of-the-art LLMs like \textit{GPT-4o}, \textit{Claude-3.7} and \textit{Gemini 2.5 Pro} .
Our experimental results show that SSA-COMET models significantly outperform AfriCOMET and are competitive with the strongest LLM (\textit{Gemini 2.5 Pro}) evaluated in our study, particularly on low-resource languages such as Twi, Luo, and \yoruba. All resources are released under open licenses to support future research.~\footnote{\href{https://huggingface.co/collections/McGill-NLP/ssa-comet-68327cdbd9300c62de8ee6c6}{Model: McGill-NLP/ssa-comet-*}}
\end{abstract}

\section{Introduction}
%
%
Recent advancements in machine translation evaluation (MTE) have largely benefited high-resource languages. Neural metrics such as COMET and MetricX~\citep{rei-etal-2020-comet, juraska-etal-2023-metricx} have demonstrated strong performance by capturing deeper semantic relationships in translations. However, their effectiveness 
diminishes for under-represented languages, such as many African languages, due to the scarcity of high-quality training and evaluation data, as well as the limitations in the multilingual large language models used as their pretrained backbones~\citep{freitag-etal-2024-llms, sai-b-etal-2023-indicmt, wang-etal-2024-evaluating}. 


\begin{figure}[t]
    \centering
    \includegraphics[width=0.5\textwidth]{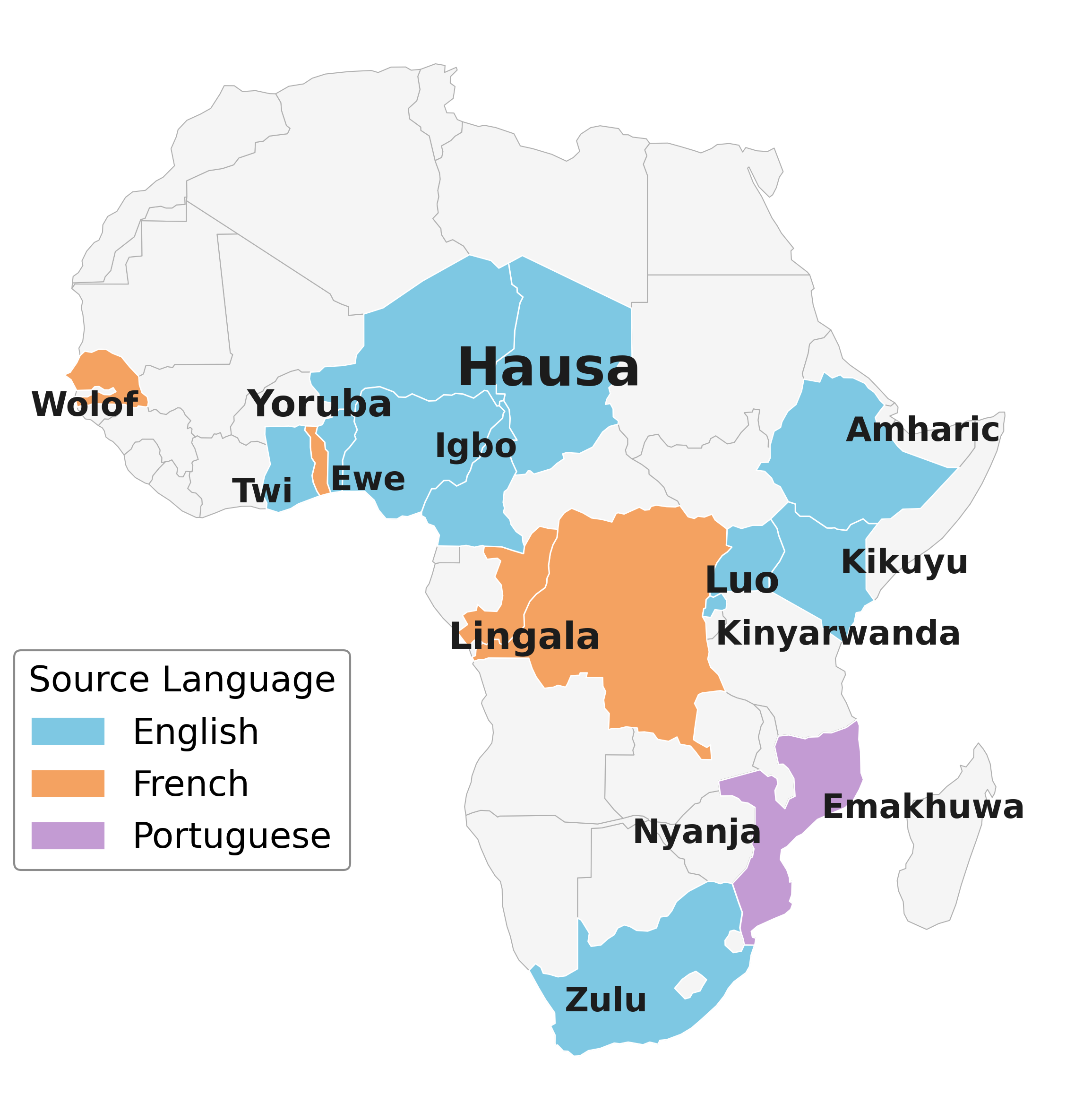}
    \vspace{-5mm}
    \caption{Language distribution across the 14 Sub-Saharan African languages in SSA-MTE.}
    \label{fig:trend-lrl-languages}
\end{figure}

To narrow this gap, \citet{wang-etal-2024-afrimte} introduced AfriMTE, a high-quality evaluation dataset covering 14 typologically diverse African languages, annotated using a simplified version of the MQM framework~\citep{Lommel2014MultidimensionalQM}, specifically designed for non-expert annotators. Building on AfriMTE, they developed AfriCOMET, an enhanced version of COMET~\citep{rei-etal-2020-comet}, by incorporating an African-centric encoder, AfroXLM-R~\citep{alabi-etal-2022-adapting}. More recently, \citet{wang-etal-2024-evaluating} enhanced these models by adopting AfroXLMR-76, which covered more African languages~\citep{adelani-etal-2024-sib}.

However, despite the advances of~\citet{wang-etal-2024-afrimte, wang-etal-2024-evaluating}, several limitations remain. First, the lack of training data in AfriMTE restricts opportunities for the broader research community to improve upon existing models. Second, the evaluation setup of AfriMTE includes only a single MT system per language pair, limiting the diversity of translation outputs and making it challenging to assess the metric’s generalizability across systems of varying quality and style. Third, the evaluation datasets in AfriMTE are relatively small—typically around 100–200 annotated examples per language pair—which may not adequately capture the full range of linguistic variation. Finally, AfriCOMET models exhibit unreliable performance for certain extreme low-resource African languages, such as Twi and Luo, producing inconsistent or low-quality estimates~\citep{adelani-etal-2025-irokobench}.

In this work, we address these challenges through three key contributions:
\textbf{(1)} We expand the landscape of high-quality MT training and evaluation data by introducing \textit{\textbf{SSA-MTE}}, a new human-annotated dataset covering 14 Sub-Saharan African language pairs,  
8 of these pairs are newly introduced compared to AfriMTE. Our annotations are sourced from the \textit{News domain}—
selected for its topical diversity, timeliness, 
and widespread use in the MT
community. The machine translated outputs in SSA-MTE are generated using a diverse set of MT systems such as Google Translate and NLLB~\citep{team2022NoLL}, and frontier large language models (LLMs) such as \textit{GPT-4o} and \textit{Gemini}. 
%
\textbf{(2)} We enhance the AfriCOMET models
by extending them on our newly collected data, resulting in \textit{\textbf{SSA-COMET}} and \textit{\textbf{SSA-COMET-QE}}, improved MTE and reference-free quality estimation (QE) metrics specifically tailored to African languages.
\textbf{(3)} We fully explore the capabilities of cutting-edge LLMs, including \textit{Gemini 2.5 Pro}, 
 \textit{GPT-4o}, 
 and \textit{Claude 3.7}, 
 for MTE and QE in a few-shot setting on the testing data of SSA-MTE.

Our experimental results demonstrate substantial overall performance improvements of the \textsc{SSA-COMET} models over \textsc{AfriCOMET}-v1.1~\citep{wang-etal-2024-evaluating}, with particularly strong gains on low-resource languages such as Twi, Luo, and \yoruba.
In MT evaluation, \textsc{SSA-COMET} demonstrates competitive performance with \textit{Gemini 2.5 Pro} and outperforms other prompting-based LLM metrics, achieving higher average Spearman correlation than \textit{GPT-4o} and \textit{Claude-3.7}, despite being an order of magnitude smaller in model size.
To support future research in African NLP and foster reproducibility, we release our dataset, models, and training pipeline under open licenses.

\section{Related Works}
Traditional MTE metrics like BLEU~\citep{papineni-etal-2002-bleu}, METEOR~\citep{banerjee-lavie-2005-meteor}, and ChrF~\citep{popovic-2015-chrf} rely on n-gram overlap and correlate poorly with human judgments. Neural metrics such as BERTScore~\citep{Zhang_020BERTScore} better capture semantic similarity. COMET~\citep{rei-etal-2020-comet} improves on this by framing MTE as a regression task using XLM-R~\citep{conneau2019unsupervised} and training data of quality scores. Its extension, COMETKiwi~\citep{rei-etal-2022-cometkiwi}, removes the need for reference translations, increasing flexibility.
More recently, MetricX~\citep{juraska-etal-2023-metricx}, which is built on mT5~\citep{xue2020mt5}, adopts a regression-based framework
similar to COMET. 
In parallel, with the rise of LLMs, there is growing interest in prompting LLMs directly to assess translation quality~\citep{kocmi2023large, freitag-etal-2024-llms}. 

Recent studies~\citep{wang-etal-2024-afrimte, wang-etal-2024-evaluating, freitag-etal-2024-llms} show that both neural metrics and prompting-based methods perform poorly on under-represented African languages, when compared to high-resource settings. To address this, AfriCOMET~\citep{wang-etal-2024-afrimte} uses an Africa-centric encoder, AfroXLMR~\citep{alabi-etal-2022-adapting}, and Non-African MTE training data to build a COMET-style metric, showing robust performance on African MTE tasks. However, recent analysis~\citep{adelani-etal-2025-irokobench} finds that AfriCOMET still shows inconsistencies with human judgments in extreme low-resource languages like Twi.

In this paper, we expand the landscape of high-quality MT training and evaluation data for African languages by introducing a newly annotated MTE dataset, and evaluate performance on newly trained COMET-based models and LLMs. 

\section{SSA-MTE: The Dataset}
This section describes the source data and MT systems used to construct SSA-MTE, presents the annotation guidelines and procedure, outlines the quality assurance measures, and provides a quantitative analysis of the resulting dataset. 

\subsection{Source Data Collection}
\paragraph{The News Domain} Given the rich structure and high quality of content in the News domain, this work focuses on the News domain, unlike AfriMTE~\citep{wang-etal-2024-afrimte}, which centers on Wikipedia data. We sourced the input sentences from the news platform \textit{Global Voices}\footnote{\url{https://globalvoices.org/}}, which publishes articles in parallel across multiple languages. Each article is tagged with topical categories such as \textit{Economics \& Business} and \textit{Education} to indicate its thematic focus. Translations on Global Voices are produced manually by a global network of volunteer contributors as part of its Lingua program, and all content is published under a Creative Commons Attribution 3.0 (CC BY 3.0) license. 

\paragraph{The Source Data} Considering that the two dominant official languages in Africa are English and French, we selected all articles available in both languages, totaling $20,419$. From this pool, we filtered for articles tagged with African regions—such as ``Guinea-Bissau'' and ``Gambia''—to ensure the content was relevant to Africa. To avoid potentially sensitive topics, we heuristically excluded articles tagged with categories such as ``war-conflict'', resulting in a subset of $3,681$ articles. Finally, we used Gemini to automatically detect and remove any remaining content that might be harmful, yielding a final collection of $1,901$ articles. From this refined set, we manually selected 200 articles by reviewing their titles and tags to ensure diverse topical coverage. At the document level, articles were segmented into sentences using the NLTK sentence tokenizer\footnote{\url{https://www.nltk.org/api/nltk.tokenize.html}}. We then applied fasttext language identification~\citep{joulin2016bag} and sentence alignment using LASER~\citep{artetxe-schwenk-2019-massively}. Sentences were retained if the language confidence score exceeded $99\%$, and sentence pairs were aligned if their similarity score was above $92.5\%$. After final deduplication, we obtained 
$1,500$ distinct parallel English–French sentence pairs for our source sentences. 

  

\paragraph{Choice of the Language Pairs (LP)}
    Given the English–French language pair, we decided to expand the coverage to 12 typologically diverse \textit{Sub-Saharan African languages}---9 using English, and 3 using French as the source language, to reflect both the Anglophone and Francophone linguistic diversity in the region. We excluded North African languages, as the most widely spoken languages in the region are Arabic dialects, which tend to yield reliable evaluation results with existing metrics such as COMET~\citep{wang-etal-2024-afrimte}. The English–sourced pairs include Amharic (\texttt{eng-amh}), Hausa (\texttt{eng-hau}), Igbo (\texttt{eng-ibo}), Kikuyu (\texttt{eng-kik}), Kinyarwanda (\texttt{eng-kin}), Luo (\texttt{eng-luo}), Twi (\texttt{egn-twi}), \yoruba (\texttt{eng-yor}), and Zulu (\texttt{eng-zul}); while the French–sourced pairs include  Ewe (\texttt{fra-ewe}), Lingala (\texttt{fra-lin}), and Wolof (\texttt{fra-wol}). Additionally, we include two low-resource Mozambique language, Emakhuwa (\texttt{vmw}) and Nyanja (\texttt{nya}), sourced from Portuguese (\texttt{por}) as detailed in Appendix~\ref{sec:appendix-por-source}. 

\subsection{MT Systems}
\label{sec:MT_sys}
To ensure a diverse representation of translation quality and styles, we used six MT systems to generate translation outputs: four closed-source models including \textit{GPT-4o}, \textit{Gemini-1.5}, \textit{Claude-3.5}
\footnote{
A template for prompting LLMs for translations is provided in Figure \ref{fig:prompt_translation}.
},
and \textit{Google Translate}, and two open-source models including \textit{NLLB-200-distilled-600M}~\citep{team2022NoLL} and \textit{M2M-100-418M}~\citep{fan2021beyond}. Since M2M-100 does not support certain languages such as Ewe and Kikuyu, we fine-tuned a separate model for each of these languages using $500,000$ randomly selected samples from the NLLB dataset\footnote{\url{https://huggingface.co/datasets/allenai/nllb}} to ensure consistent translation quality. During this procedure, Kikuyu was not supported by Google Translate; therefore, translations for this language were generated using only five systems. Similarly, for Ewe and Wolof, we excluded GPT-4o outputs, as the model declined to produce translations in more than half of the cases. The MT outputs for \texttt{por-vmw} are detailed in Appendix~\ref{sec:appendix-MT-vma}.

\subsection{Annotation Guidelines, Tool, and Protocol}
Building on the success of the simplified MQM annotation guidelines proposed by~\citet{wang-etal-2024-afrimte}, we adopt the same framework for both error-span and scoring annotations in this work. Specifically, we evaluate the adequacy of each machine translation output. Evaluators review both the source and translated texts, highlighting error spans, categorized as ``\texttt{Addition}'', ``\texttt{Omission}'', ``\texttt{Mistranslation}'', and ``\texttt{Untranslated}''. They then assign an overall translation quality score using a continuous direct assessment (DA) scale ranging from 0 to 100, 
strictly following the annotation protocol established in~\citet{wang-etal-2024-afrimte}.

We used the same annotation tool introduced in~\citet{wang-etal-2024-afrimte},\footnote{\url{https://github.com/marek357/annotation-tool-frontend}} which provides an interface supporting both error span highlighting and DA scoring,
and allows each evaluator to work independently. For each LP, we recruited \textit{two bilingual native speakers} with at least a Bachelor's degree to serve as evaluators. Annotation work was evenly divided, with 300 overlapping samples included to assess inter-evaluator agreement for quality assurance. Reference translations per LP were produced by two professional translators, who manually translated the sources from scratch, without using any machine translation tools.
We annotated 6,600 samples per language pair, including 300 overlapping samples for inter-evaluator agreement, and
this results in \textit{6,300 distinct samples} per LP, evenly distributed across MT systems.\footnote{
For languages not supported by certain MT systems, annotations were distributed across five systems instead of six.}
%
For each LP, all 1,500 source sentences were translated into the target African language. 
%

\subsection{Annotation Quality Assurance}
We employed several measures to assure the quality of the annotated data.

\paragraph{Evaluator Selection} To select qualified evaluators from a candidates pool, we followed the training procedure outlined in~\citet{wang-etal-2024-afrimte}. Each candidate was required to complete an annotation test designed to both familiarize them with the annotation tool and evaluate their understanding of the annotation guidelines. The test included 22 samples: 20 unique samples drawn from the dataset and 2 repeated samples to assess self-consistency.
We assessed the submitted annotations using a heuristic quality check. Specifically, we flagged cases where the assigned score and the highlighted error spans were inconsistent—for example, when a score below 80 was assigned without any error spans, or when a score of 100 was given despite the presence of errors. Moreover,
Inter-evaluator agreement was measured by checking whether score differences were below 20 among evaluators. For the repeated samples, we evaluated each candidate’s self-consistency, defined as producing similar error spans and assigning scores that differed by less than 5. Finally, a manual review was conducted to ensure overall annotation quality.
For each LP, we select the top two evaluators who satisfied four criteria: (1) more than 80\% agreement with each other, (2) minimal heuristic quality issues, (3) high self-consistency, and (4) a satisfactory outcome in manual quality review.


\begin{figure*}[t]
\begin{center}
\vskip -0.5cm
\includegraphics[width=0.9\linewidth]{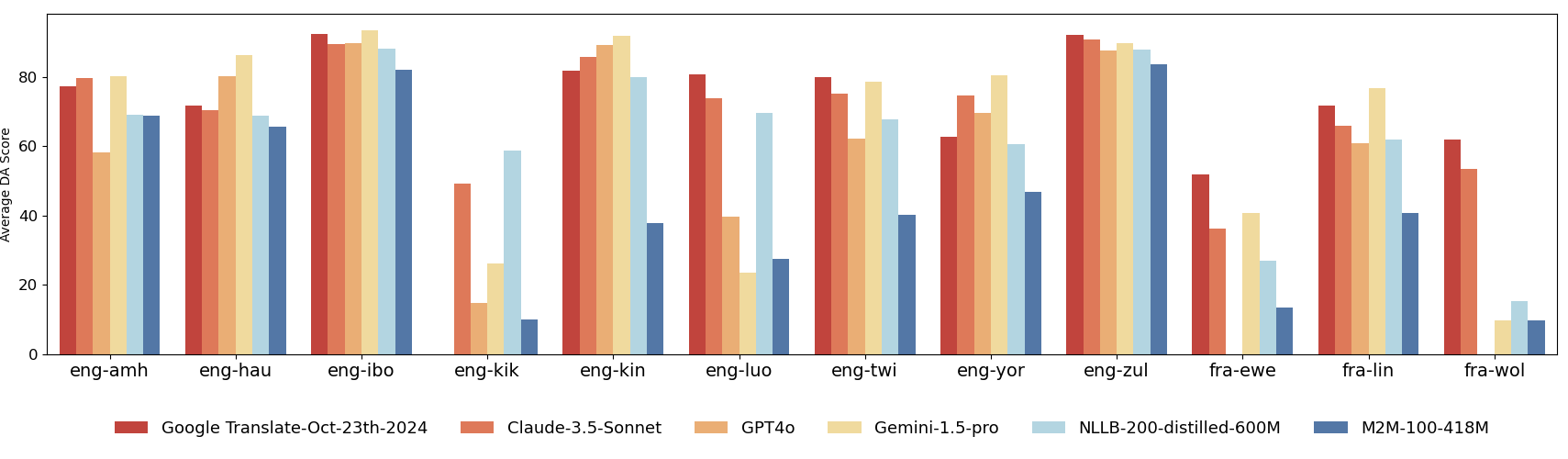} 
\end{center}
\vskip -0.4cm
\caption{\textbf{Average DA scores across MT systems and LPs}. Low-resource pairs such as \texttt{eng-kik} and \texttt{fra-wol} remain particularly challenging for current translation systems.}
\label{fig:systems_scores}

\end{figure*}
\paragraph{Agreement on the Overlaps} 
After selecting the evaluators, we implemented a quality assurance procedure using annotations on the 300 overlapping samples. These samples were independently annotated by both evaluators and served to assess inter-evaluator agreement. To evaluate annotation quality and consistency, we computed Spearman-rank and Pearson correlation coefficients, as well as the Intraclass Correlation Coefficient (ICC) between their assigned scores. Since the evaluators were fixed for each language pair (i.e., the only raters of interest), we used the two-way mixed-effects model ICC(3,,$k$), with $k = 2$ in our setup.

\begin{table}[t]
\centering
\small
\begin{tabular}{l|c|c|c}
\toprule
\textbf{LP} & \textbf{Pearson} & \textbf{Spearman} & \textbf{ICC(3,2)} \\
\midrule
\texttt{eng-amh} & 0.597 & 0.653 & 0.747 \\
\texttt{eng-hau} & 0.406 & 0.476 & 0.573 \\
\texttt{eng-ibo} & 0.314 & 0.253 & 0.358 \\
\texttt{eng-kik} & 0.735 & 0.776 & 0.847 \\
\texttt{eng-kin} & 0.486 & 0.513 & 0.632 \\
\texttt{eng-luo} & 0.735 & 0.724 & 0.842 \\
\texttt{eng-twi} & 0.757 & 0.772 & 0.862 \\
\texttt{eng-yor} & 0.567 & 0.520 & 0.723 \\
\texttt{eng-zul} & 0.249 & 0.107 & 0.392 \\
\texttt{fra-ewe} & 0.560 & 0.612 & 0.694 \\
\texttt{fra-lin} & 0.399 & 0.339 & 0.570 \\
\texttt{fra-wol} & 0.592 & 0.648 & 0.741 \\
\texttt{por-vmw} & 0.620 & 0.580 & 0.764 \\
\texttt{por-nya} & 0.812 & 0.751 & 0.896 \\
\hline
\end{tabular}
\vspace{-2mm}
\caption{Inter-annotator agreement metrics (Pearson, Spearman-rank, ICC(3,2) on 300 overlapping samples.}
\label{tab:correlation_icc}
\end{table}

 To reduce evaluator bias, we first normalized the DA scores at the evaluator level, converting them to z-scores. We then computed the agreement statistics described above on the \textit{300 overlapping samples}. The results are presented in Table~\ref{tab:correlation_icc}. LPs that exhibited at least a moderate level of agreement, defined as having \textit{both Spearman rank and Pearson correlation coefficients above $0.4$ and an ICC value above $0.5$}—were included in the training, development, and test sets. 
%
As a result, 11 LPs were selected for inclusion in all three data splits: \texttt{eng-amh}, \texttt{eng-hau}, \texttt{eng-kik}, \texttt{eng-kin}, \texttt{eng-luo}, \texttt{eng-twi}, \texttt{eng-yor}, \texttt{fra-ewe}, \texttt{fra-wol},  \texttt{por-vmw} and  \texttt{por-nya} .
Although the remaining LPs did not meet the criteria, we retained them for training to introduce additional language diversity, which may help improve the robustness and generalization for modeling.


Among the three remaining LPs (\texttt{eng-ibo}, \texttt{eng-zul}, and \texttt{fra-lin}), \texttt{fra-lin} showed a Pearson correlation close to $0.4$ and an ICC above $0.5$, indicating moderate positive correlation and agreement, though its Spearman rank correlation was slightly lower at $0.339$. Given its relatively acceptable agreement levels, we included \texttt{fra-lin} in the training data without additional filtering. 
In contrast, for \texttt{eng-ibo} and \texttt{eng-zul}, which exhibited weaker agreement across all metrics, we applied further filtering to remove low-quality annotations before including them in training. The detailed filtering process is described in Appendix~\ref{appedix-further-selection}.

\begin{table}[t]
\centering
\footnotesize
\begin{tabular}{lccc}
\toprule
\textbf{LP} & \textbf{Train} & \textbf{Dev} & \textbf{Test} \\
\midrule
\texttt{eng-amh} & 4563 & 326 & 1166 \\
\texttt{eng-hau} & 4693 & 338 & 1192 \\
\texttt{eng-ibo} & 1501 & --  & --   \\
\texttt{eng-kik} & 4752 & 318 & 1172 \\
\texttt{eng-kin} & 4768 & 349 & 1210 \\
\texttt{eng-luo} & 4691 & 341 & 1199 \\
\texttt{eng-twi} & 4820 & 325 & 1200 \\
\texttt{eng-yor} & 4717 & 333 & 1206 \\
\texttt{eng-zul} & 1905 & --  & --   \\
\texttt{fra-ewe} & 4423 & 296 & 1077 \\
\texttt{fra-lin} & 4626 & -- & -- \\
\texttt{fra-wol} & 4874 & 341 & 1175 \\
\texttt{por-vmw}  &3309 &	130 &	930\\
\texttt{por-nya} & 3489&	135	&1241 \\
\midrule
\textbf{Total} &57131&	3232&	12768\\
\bottomrule
\end{tabular}
\caption{Number of \textbf{training}, \textbf{development}, and \textbf{test} examples in SSA-MTE for each LP.}
\label{tab:split_sizes}
\vspace{-3mm}
\end{table}

To introduce further diversity in evaluation, we include two previously under-studied LPs—\textit{\textbf{Portuguese to Emakhuwa (vmw)}} and \textit{\textbf{Portuguese to Nyanja (nya)}}, 
from Mozambique.
Follows the same design as the other 12 LPs: focusing on the \textit{news domain}, \textit{multiple MT systems} are included, and the same \textit{ annotation and quality assurance procedures} are applied.
Details of the data collection and MT generation processes are provided in Appendix~\ref{sec:appendix-por-source}, ~\ref{sec:appendix-MT-vma}, and~\ref{sec:appendix-process-vma}.

\subsection{Final Data Statistics}
For the final version of the dataset, we applied several filtering steps to ensure high-quality annotations. First, we \textit{excluded all cases with a score below 80} that lacked annotated error spans. We also removed cases falling in the top 20\% of DA scores but within the bottom 20\% of ChrF scores relative to the reference translations. Similarly, we filtered out cases in the bottom 20\% of DA scores that had the highest 20\% of ChrF scores. 

To avoid potential information leakage, \textbf{DEV} and \textbf{TEST} sets were selected based on source documents: we \textit{excluded the 300 overlapping examples} used for inter-evaluator agreement and randomly sampled \textit{40 source documents} for the TEST set and \textit{10 documents} for the DEV set. For all languages, only translations whose source sentences came from these selected documents were included in the DEV and TEST sets. This document-level selection helps prevent models from learning translation patterns from highly similar source texts. The \textit{remaining data} was assigned to the \textbf{TRAIN} set. Final dataset statistics are reported in Table \ref{tab:split_sizes}.

To view the translation quality of each MT system for each LP, we present the average DA scores across LPs and MT systems in Figure~\ref{fig:systems_scores}. High-resource LPs, such as \texttt{English–Zulu}, generally achieve higher scores, whereas low-resource pairs like \texttt{English–Kikuyu} and \texttt{French–Wolof} exhibit substantially lower translation quality.





\section{SSA-COMET Models}
In this section, we describe the modeling approaches of SSA-COMET and SSA-COMET-QE. 

\subsection{Modeling Methods}

\paragraph{MTE Modeling} We follow the modeling setup the same as \textbf{COMET} for developing MTE systems for African languages. 
Our models are trained to predict DA adequacy scores, using the \textbf{COMET} architecture, which is based on a regression-based estimator framework. We implement both single-task learning (STL) and multi-task learning (MTL). 

\paragraph{Single-Task Learning (STL)} In the STL setting, each of the source (\texttt{src}), machine translation (\texttt{mt}), and reference (\texttt{ref}) segments is independently encoded using a multilingual encoder. The resulting sentence embeddings are pooled, concatenated, and passed through a feed-forward regressor trained to minimize mean squared error against the human-annotated adequacy scores.

\paragraph{Multi-Task Learning (MTL)} In the MTL setting, we adopt the unified multi-view formulation from \citealp{wan2022unite}, where the model is trained \textbf{jointly} on three input configurations: \texttt{⟨src, mt⟩}, \texttt{⟨mt, ref⟩}, and \texttt{⟨src, mt, ref⟩}. Each configuration is passed through the model to produce a separate prediction, and the final score is computed by averaging the three outputs. This formulation leverages multiple input perspectives to provide richer supervision and improve generalization. 

\paragraph{QE Modeling} Additionally, we develop SSA-COMET-QE, a variant that mirrors the 
AfriCOMET-QE
architecture. This model operates solely on the \texttt{⟨src, mt⟩} pair and is optimized for the QE setting. It is trained independently using the same DA scores, enabling direct quality estimation without relying on reference translations.

\subsection{African-centric Multilingual Encoder}
\citet{wang-etal-2024-afrimte} demonstrated that employing an African-centric encoder, AfroXLMR~\citep{alabi-etal-2022-adapting}, trained on 17 languages, leads to improved machine translation evaluation (MTE) for African languages. However, its performance deteriorates for languages outside its training coverage. The variant with broader language coverage, AfroXLMR-76L, yielded further improvements for languages such as Twi and Wolof. In our SSA-MTE experiments, we observed that AfroXLMR-76L~\cite{adelani-etal-2024-sib} does not support Emakhuwa, resulting in inferior performance (see \S\ref{sec:ablation_afroxlmr}). To address this limitation, we adopted multilingual adaptive fine-tuning, extending the strategy of AfroXLMR-76L by continuing pre-training on both large monolingual corpora (for languages with at least 1 MB of data) and machine-translated corpora generated from NLLB-200 (for languages with less than 10MB and \yoruba language). In addition, we incorporated the recently released parallel Emakhuwa–Portuguese corpus~\cite{ali-etal-2024-expanding} into the pre-training process. The resulting model, AfroXLMR-114L,\footnote{We release the model on HuggingFace \url{https://huggingface.co/Davlan/afro-xlmr-large-114L}} now covers 110 African languages alongside four high-resource languages--English, French, Arabic, and Portuguese, that are widely spoken in Africa. 




\section{Experiment Setup}
For the TEST evaluation, to ensure comparability across language pairs and annotators, all human-annotated DA scores in the test set were first standardized using $z$-score normalization.

\textbf{SSA-COMET training} We combine the training data used for AfriCOMET (the WMT Non-African DA data) with the training split of the newly annotated SSA-MTE. Score pre-processing is conducted in two steps: we first apply $z$-normalization at the evaluator level, followed by min-max scaling to improve consistency and interpretability. To establish a stable global range, we collect the 800 highest and 800 lowest $z$-scores across all languages and use their corresponding averages to define the minimum and maximum values. The resulting scores are then scaled and clipped to fall within the [0, 1] range. The DEV sets from both AfriMTE and our new dataset are used as validation data during training.

\textbf{LLM-based evaluation} We sample the few-shot examples from the training split of the SSA-MTE dataset. For the \texttt{por-vmw} language pair, which does not have a training split, demonstrations were instead sampled from the processed and filtered 300 overlapping annotated examples used to assess inter-annotator agreement.
\subsection{Model Configurations}
We use the multilingual encoder AfroXLMR-114L, pretrained on 114 languages widely spoken in Africa. All models are trained using the open-source COMET codebase. Training for the STL and QE models is conducted on a single NVIDIA L40S GPU, while the MTL model is trained on a single NVIDIA A100-SXM4-80GB GPU. We use a batch size of 16 with gradient accumulation over 2 steps. All other hyperparameters follow the default configuration used in AfriCOMETv1.1.

\begin{table*}[htbp]
\centering
\large
\setlength\tabcolsep{4pt}
\resizebox{\textwidth}{!}{%
\begin{tabular}{c|cc|ccccc|cc|cc}
\toprule
\makecell[l]{LP} & \textbf{Bleu} & \textbf{ChrF++} & 
\textbf{COMET22} & 
\makecell[c]{\textbf{AfriCOMET} \\ \textbf{v1.1 STL}} & 
\makecell[c]{\textbf{AfriCOMET} \\ \textbf{v1.0 MTL}} & 
\makecell[c]{\textbf{AfriCOMET} \\ \textbf{v1.1 MTL}} & 
\textbf{MetricX 24} & 
\textbf{Claude-3.7} & 
\makecell[c]{\textbf{Gemini-pro} \\ \textbf{2.5}} & 
\makecell[c]{\textbf{SSA-COMET} \\ \textbf{STL}} & 
\makecell[c]{\textbf{SSA-COMET} \\ \textbf{MTL}} \\
\midrule
\texttt{eng-amh} & 0.352 & 0.441 & 0.548 & 0.588 & 0.612 & 0.604 & \textbf{0.659} & 0.566 & 0.605 & 0.570 & 0.615 \\
\texttt{eng-hau} & 0.312 & 0.402 & 0.405 & 0.465 & 0.479 & 0.476 & 0.495 & 0.425 & 0.471 & 0.463 & \textbf{0.502} \\
\texttt{eng-kik} & 0.505 & 0.599 & 0.263 & 0.492 & 0.556 & 0.693 & 0.622 & 0.696 & 0.735 & 0.707 & \textbf{0.764} \\
\texttt{eng-kin} & 0.392 & 0.459 & 0.335 & 0.507 & 0.551 & 0.532 & \textbf{0.620} & 0.536 & 0.528 & 0.568 & 0.584 \\
\texttt{eng-luo} & 0.465 & 0.612 & 0.361 & 0.616 & 0.496 & 0.693 & 0.543 & 0.678 & \textbf{0.782} & 0.667 & 0.773 \\
\texttt{eng-twi} & 0.364 & 0.502 & 0.328 & 0.527 & 0.537 & 0.596 & 0.637 & 0.652 & \textbf{0.710} & 0.624 & 0.687 \\
\texttt{eng-yor} & 0.382 & 0.436 & 0.349 & 0.442 & 0.482 & 0.476 & 0.455 & 0.501 & 0.524 & 0.545 & \textbf{0.604} \\
\texttt{fra-ewe} & 0.311 & 0.426 & 0.330 & 0.443 & 0.494 & 0.550 & 0.581 & 0.614 & \textbf{0.658} & 0.565 & 0.618 \\
\texttt{fra-wol} & 0.476 & 0.572 & 0.304 & 0.493 & 0.478 & 0.518 & 0.560 & 0.699 & \textbf{0.750} & 0.661 & 0.724 \\
\texttt{por-vmw} & 0.181 & 0.414 & 0.198 & 0.238 & 0.277 & 0.237 & 0.378 & 0.463 & \textbf{0.487} & 0.395 & 0.454 \\
\texttt{por-nya} &0.518 &	0.613&	0.607	&0.685	&0.680&	0.677&	0.686&	0.684	&\textbf{0.709}	&0.679&	0.704 \\
\midrule
Average & 0.387 &	0.498&	0.366	&0.500&	0.513&	0.550	&0.567&	0.592	&0.633	&0.586	&\textbf{0.639} \\
\bottomrule
\end{tabular}
}
\vspace{-2mm}
\caption{\textbf{Spearman correlation of MTE metrics with human judgments across LPs}. The best scores are \textbf{bolded}.}
\label{tab:mte_spearman_3dp}
\end{table*}

\begin{table*}[htbp]
\centering
\scriptsize
\setlength\tabcolsep{4pt}
\resizebox{\textwidth}{!}{%
\begin{tabular}{l|cc|cc|cc|cc|cc|cc}
\toprule
\multirow{2}{*}{\textbf{LP}}
& \multicolumn{2}{c|}{\textbf{AfriCOMETv1.1-MTL}} 
& \multicolumn{2}{c|}{\textbf{MetricX-24}} 
& \multicolumn{2}{c|}{\textbf{Claude-3.7-Sonnet}} 
& \multicolumn{2}{c|}{\textbf{Gemini-2.5 Pro}} 
& \multicolumn{2}{c|}{\textbf{SSA-COMET-QE}} 
& \multicolumn{2}{c}{\textbf{SSA-COMET-MTL}} \\
\cmidrule(lr){2-3} \cmidrule(lr){4-5}
\cmidrule(lr){6-7} \cmidrule(lr){8-9}
\cmidrule(lr){10-11} \cmidrule(lr){12-13}
& \textbf{Spear.} & \textbf{Pear.} 
& \textbf{Spear.} & \textbf{Pear.} 
& \textbf{Spear.} & \textbf{Pear.} 
& \textbf{Spear.} & \textbf{Pear.} 
& \textbf{Spear.} & \textbf{Pear.} 
& \textbf{Spear.} & \textbf{Pear.} \\
\midrule
\texttt{eng-amh} & 0.568 & 0.619 & \textbf{0.618} & \textbf{0.639} & 0.541 & 0.587 & 0.558 & 0.593 & 0.537 & 0.573 & 0.556 & 0.592 \\
\texttt{eng-hau} & 0.388 & 0.405 & \textbf{0.436} & 0.416 & 0.357 & 0.382 & 0.401 & 0.412 & 0.375 & 0.392 & 0.414 & \textbf{0.432} \\
\texttt{eng-kik} & 0.655 & 0.648 & 0.464 & 0.452 & 0.677 & 0.609 & 0.703 & 0.650 & 0.685 & 0.647 & \textbf{0.733} & \textbf{0.732} \\
\texttt{eng-kin} & 0.473 & 0.619 & \textbf{0.592} & 0.738 & 0.530 & 0.694 & 0.534 & 0.714 & 0.530 & 0.737 & 0.546 & \textbf{0.775} \\
\texttt{eng-luo} & 0.644 & 0.638 & 0.329 & 0.332 & 0.672 & 0.646 & \textbf{0.757} & 0.721 & 0.646 & 0.636 & 0.739 & \textbf{0.738} \\
\texttt{eng-twi} & 0.561 & 0.678 & 0.563 & 0.686 & 0.640 & 0.747 & \textbf{0.697} & \textbf{0.771} & 0.617 & 0.708 & 0.635 & 0.738 \\
\texttt{eng-yor} & 0.424 & 0.501 & 0.405 & 0.524 & 0.492 & 0.595 & 0.531 & 0.607 & 0.489 & 0.580 & \textbf{0.581} &\textbf{0.651} \\
\texttt{fra-ewe} & 0.483 & 0.437 & 0.476 & 0.430 & 0.592 & 0.518 & \textbf{0.623} & 0.524 & 0.501 & 0.480 & 0.576 & \textbf{0.550} \\
\texttt{fra-wol} & 0.407 & 0.358 & 0.291 & 0.258 & 0.687 & 0.623 & \textbf{0.743} & \textbf{0.676} & 0.614 & 0.545 & 0.676 & 0.637 \\
\texttt{por-vmw} & 0.134 & 0.168 & 0.292 & 0.369 & \textbf{0.498} & \textbf{0.551} & 0.481 & 0.528 & 0.343 & 0.417 & 0.403 & 0.481 \\
\texttt{por-nya} & 0.657 & 0.885 & 0.673 & 0.931 & 0.688 & 0.887 & 0.678 & 0.895 & 0.678 & 0.904 & \textbf{0.692} & \textbf{0.918} \\
\midrule
\textbf{Average} & 0.490 & 0.542 & 0.467 & 0.525 & 0.569 & 0.595 & \textbf{0.610} & 0.645 & 0.547 & 0.602 & 0.595 & \textbf{0.659} \\
\bottomrule
\end{tabular}
}
\vspace{-2mm}
\caption{\textbf{QE results (Spearman and Pearson correlations) for each LP}. The best scores are \textbf{bolded}. }
\label{tab:qe-spearman-pearson}
\end{table*}

\subsection{Baselines}
\label{sec:baselines}

To benchmark the performance of SSA-COMET, we compare it against a wide range of baselines across both MTE and QE settings. These include:

\paragraph{Traditional metrics for MTE}
\textit{BLEU} and \textit{ChrF++} are lexical overlap metrics based on n-gram precision and character-level F-scores, respectively. 

\paragraph{Neural regression-based metrics for MTE}
For evaluation under the MTE setting, we include \textit{COMET22}, \textit{MetricX-24}, \textit{AfriCOMETv1.0-MTL}~\citep{wang-etal-2024-afrimte} (based on AfroXLMR that supports 20 African languages), \textit{AfriCOMETv1.1-STL}~\citep{wang-etal-2024-evaluating} (based on AfroXLMR-76L supporting 76 languages), and \textit{AfriCOMETv1.1-MTL}.
The latter is a self-replication model, trained on the same data as \textit{AfriCOMET v1.1-STL} but using a multi-task learning formulation.

\paragraph{Neural regression-based metrics for QE}
For the QE setting, we evaluate \textit{MetricX-24} and \textit{AfriCOMET v1.1-MTL} in QE mode by disabling the reference input at inference time.

\paragraph{LLM baselines}
We evaluate four open-weight LLMs such as \textit{Gemma-3 27B-it}, \textit{LLaMA-4 100B}, \textit{LLaMA-4 400B}, and \textit{DeepSeek V3}. Additionally, we conducted an evaluation using some frontier proprietary models such as \textit{GPT-4o (08/24)}, and \textit{Gemini-2.0 Flash}, \textit{Claude-3.7-Sonnet} and \textit{Gemini-2.5 Pro} under both MTE and QE settings as strong prompting-based baselines. 

We adopt a \textbf{5-shot prompt} setup, guided by the same annotation instructions provided to human annotators. To ensure broad coverage of translation quality levels, we extract the minimum and maximum adequacy scores from the training set and divide the range into five equal intervals. One example is sampled from each interval to construct the 5-shot prompt. The same set of demonstrations is used across all test cases for each language pair to ensure consistency and fairness in evaluation.
We experiment with two prompting templates: one that includes error span detection before adequacy scoring, and one that directly predicts the score without error identification. Full templates for both setups are provided in Figure~\ref{fig:template_with_error_span} and Figure~\ref{fig:llm-prompt-wo-MTE}.

\begin{table*}[htbp]
\centering
\scriptsize
\setlength\tabcolsep{4pt}
\resizebox{\textwidth}{!}{%
\begin{tabular}{l|c|cccc|cccc}
\toprule
 \multirow{2}{*}{\textbf{Metric}}& \textbf{w/ Error} 
& \textbf{Gemma 3} & \textbf{Llama 4} & \textbf{Llama 4} 
& \textbf{Deepseek V3} & \textbf{GPT-4o} & \textbf{Gemini-2.0} 
& \textbf{Claude-3.7} & \textbf{Gemini-2.5} \\
 & \textbf{Span} & \textbf{27B} & \textbf{100B} & \textbf{400B} 
& \textbf{671B} & \textbf{(Aug-2024)} & \textbf{Flash} 
& \textbf{Sonnet} & \textbf{Pro} \\
\midrule
\multirow{2}{*}{Spearman} 
& $\times$  & 0.470 & 0.463 & 0.525 & 0.514 & 0.519 & 0.556 & 0.592 & \textbf{0.633} \\
& \checkmark & 0.365 & 0.298 & 0.499 & 0.354 & 0.361 & 0.521 & 0.585 & \textbf{0.595} \\
\midrule
\multirow{2}{*}{Pearson} 
& $\times$ & 0.535 & 0.519 & 0.581 & 0.560 & 0.575 & 0.606 & 0.634 & \textbf{0.667} \\
& \checkmark & 0.415 & 0.322 & 0.540 & 0.382 & 0.402 & 0.555 & 0.632 & \textbf{0.644} \\
\bottomrule
\end{tabular}
}
\vspace{-3mm}
\caption{\textbf{Average correlation performance of LLMs (Spearman and Pearson) across all LPs}, with and without error span annotation prompts. The best scores are \textbf{bolded}.}
\label{tab:error-span-results}
\end{table*}

\begin{table}[htbp]
\centering
\resizebox{\columnwidth}{!}{%
\begin{tabular}{l|cc|cc}
\toprule
\multirow{2}{*}{\textbf{LP}} 
& \multicolumn{2}{c|}{\textbf{SSA-COMET-STL}} 
& \multicolumn{2}{c}{\textbf{SSA-COMET-MTL}} \\
\cmidrule(lr){2-3} \cmidrule(lr){4-5}
& \textbf{w/ WMT}  & \textbf{w/o WMT}
& \textbf{w/ WMT}  & \textbf{w/o WMT}
\\
\midrule
\texttt{eng-amh}  & \textbf{0.570} & 0.546 & \textbf{0.615} & 0.556 \\
\texttt{eng-hau}  & \textbf{0.463} & 0.435 & \textbf{0.502} & 0.459 \\
\texttt{eng-kik}  & \textbf{0.707} & 0.701 & \textbf{0.764} & 0.739 \\
\texttt{eng-kin}  & 0.568 & 0.568 & \textbf{0.584} & 0.569 \\
\texttt{eng-luo}  & \textbf{0.667} & 0.662 & \textbf{0.773} & 0.746 \\
\texttt{eng-twi}  & \textbf{0.624} & 0.621 & \textbf{0.687} & 0.654 \\
\texttt{eng-yor}  & \textbf{0.545} & 0.520 & \textbf{0.604} & 0.558 \\
\texttt{fra-ewe}  & 0.565 & 0.565 & \textbf{0.618} & 0.579 \\
\texttt{fra-wol}  & 0.661 & \textbf{0.677} & \textbf{0.724} & 0.720 \\
\texttt{por-vmw}  & \textbf{0.395} & 0.365 & \textbf{0.454} & 0.422 \\
\texttt{por-nya}  & 0.679 & \textbf{0.684} & \textbf{0.704} & 0.694 \\
\midrule
\textbf{Average} 
         & \textbf{0.586} & 0.577 & \textbf{0.639} & 0.609 \\
\bottomrule
\end{tabular}
}
\vspace{-3mm}
\caption{\textbf{Spearman correlations for SSA-COMET in STL and MTL setting}--trained with and without WMT data. The best scores are \textbf{bolded}.}
\label{tab:spearman-wmt}
\end{table}

\subsection{Main Findings}
\paragraph{Superior performance of SSA-COMET in MTE}
As shown in Table~\ref{tab:mte_spearman_3dp}, SSA-COMET-MTL achieves the second highest average Spearman correlation with human judgments in the MTE setting, outperforming all prior AfriCOMET variants as well as the strong prompting-based baselines such as Gemini-2.5 Pro.

\paragraph{Robust QE performance}
A similar trend is observed in Table \ref{tab:qe-spearman-pearson}. Under QE setting, SSA-COMET-MTL ranks first in terms of Pearson correlation and second in Spearman correlation. When excluding the por-vmw and por-nya language pairs, SSA-COMET-MTL achieves the highest average performance across the remaining language pairs.

\paragraph{Gains in Previously Challenging Low-Resource Languages}
Notably, SSA-COMET shows remarkable improvements on low-resource language pairs where all previous AfriCOMET variants have consistently struggled—particularly on Twi and Wolof. As shown in both Table~\ref{tab:mte_spearman_3dp} and Table~\ref{tab:qe-spearman-pearson}, our model achieves substantial gains in correlation with human judgments for these languages. These results highlight the critical role of in-language, high-quality training data, which allows the model to better capture language-specific characteristics and produce more accurate and reliable quality estimates in low-resource scenarios.



\paragraph{LLM-based prompting is more Robust to the absence of Reference}
LLMs demonstrate greater robustness to the absence of reference translations. Regression-based metrics achieved worse performance when 
changing
from MTE to QE settings. As shown in Table~\ref{tab:mte_spearman_3dp} and Table~\ref{tab:qe-spearman-pearson}, the drop in Spearman correlation from MTE to QE is relatively small for Claude-3.7 (0.022 on average) and Gemini-2.5 Pro (0.023 on average), in contrast to the 
obvious declines observed in regression-based models.
This indicates that regression models are more dependent on the presence of reference translations compared to LLMs.
Despite the impressive LLM performance, their performance is significantly worse results if we do not provide in-context examples (5-shots) as shown in Appendix~\ref{sec:zero-shot-prompting}. 

\paragraph{Impact of Error Span Prediction on LLMs}
Table~\ref{tab:error-span-results} presents a comparison of LLM performance with and without error span prediction. We observe a consistent decline in both Spearman and Pearson correlations when models are prompted to identify error spans prior to generating adequacy scores. For example, Gemini-2.5 Pro’s Spearman correlation drops from 0.633 to 0.595, and its Pearson correlation decreases from 0.667 to 0.633. Overall, prompting for error spans before generating the final score does not appear to improve the quality of final predictions. We provide some \textbf{qualitative analysis} for \yoruba showing that the predicted spans are often reliable in Appendix~\ref{sec:qualitative}. Further investigation is still needed to show how useful the predictions are to users of various MT systems. 

\subsection{Ablation: Impact of WMT Data}
Table~\ref{tab:spearman-wmt} presents a performance comparison of models trained with and without WMT Non-African data augmentation. As shown, incorporating WMT data yields notable gains in the MTL setting, whereas its impact in the STL setting is comparatively limited. Notably, our annotated SSA-MTE dataset proves highly effective: the model trained solely on SSA-MTE achieves an average Spearman correlation of $0.586$ under the MTL setup, already outperforming all AfriCOMET baselines (as shown in Table~\ref{tab:mte_spearman_3dp}). This highlights the quality and utility of our in-domain annotations, demonstrating that strong performance can be attained even without external training data.

\subsection{Ablation: AfroXLMR-114L vs. AfroXLMR-76L}
\label{sec:ablation_afroxlmr}
To examine the effect of the base model choice, we finetuned a variant of SSA-COMET on AfroXLMR-76L, which is the same base model used in AfriCOMET. Unlike AfroXLMR-114L, the 76L model does not include Emakhuwa in its pretraining data. This setup allows us to directly assess whether the broader language coverage of AfroXLMR-114L contributes to stronger correlations.

As shown in Table~\ref{tab:114vs76}, SSA-COMET trained on AfroXLMR-114L achieves higher overall correlations than the 76L counterpart, with particularly large gains on the \texttt{por-vmw} language pair, which is absent from the 76L pretraining corpus. In addition, thanks to the inclusion of more \yoruba data, we also observe a clear improvement on the \texttt{eng-yor} language pair. These results demonstrate that extended pretraining coverage leads to more robust evaluation performance in low-resource African languages.

\subsection{Ablation: With/Without filtered data from excluded languages}

We initially annotated data for three additional languages: Igbo, Swahili, and Sesotho. However, the Spearman correlations on the 300 overlapping portions were below 0.20, leading us to exclude them from the training data without special handling. Instead, we relied on three silver labellers: AfriCOMET-v1.1-MTL, MetricX-24, and the silver labellers described in Appendix~\ref{appedix-further-selection}. We selected the data combination that maintained high correlations across all three silver labellers. The selection was performed greedily: at each step, we removed the sample that most improved the overall Spearman correlations, continuing until the correlations exceeded the threshold of 0.5.

As we can observe in Table \ref{tab:114vs76}, adding the filtered data yields slight overall gains, especially for AfroXLMR-114L (0.639 → 0.642). This shows that carefully filtered data provides complementary supervision that generalizes well. We will release all four MTE models 
on HuggingFace. 

\begin{table}[t]
\centering
\scriptsize
\resizebox{0.9\columnwidth}{!}{%
\begin{tabular}{l|cc|cc}
\toprule
\multirow{2}{*}{\textbf{LP}} 
& \multicolumn{2}{c|}{\textbf{w/o excl. lang.} } 
& \multicolumn{2}{c}{\textbf{w/ excl. lang. }} \\
\cmidrule(lr){2-3} \cmidrule(lr){4-5}
& \textbf{114L}  & \textbf{76L}
& \textbf{114L}  & \textbf{76L}
\\
\midrule
\texttt{eng-amh}  & 0.615 & \textbf{0.633} & 0.609 & 0.630 \\
\texttt{eng-hau}  & 0.502 & 0.499 & \textbf{0.507} & 0.500 \\
\texttt{eng-kik}  & \textbf{0.764} & 0.763 & 0.762 & 0.763 \\
\texttt{eng-kin}  & 0.584 & \textbf{0.607} & 0.590 & 0.602 \\
\texttt{eng-luo}  & \textbf{0.773} & 0.771 & 0.770 & 0.765 \\
\texttt{eng-twi}  & 0.687 & \textbf{0.703} & 0.686 & 0.698 \\
\texttt{eng-yor}  & \textbf{0.604} & 0.557 & 0.591 & 0.554 \\
\texttt{fra-ewe}  & 0.618 & 0.637 & 0.654 & \textbf{0.659} \\
\texttt{fra-wol}  & 0.724 & 0.722 & \textbf{0.726} & 0.724 \\
\texttt{por-vmw}  & 0.454 & 0.427 & \textbf{0.459} & 0.424 \\
\texttt{por-nya}  & 0.704 & 0.700 & \textbf{0.706 }& 0.703 \\
\midrule
\textbf{Average} 
         & 0.639 & 0.638 & \textbf{0.642} & 0.639 \\
\bottomrule
\end{tabular}
}
\vspace{-3mm}
\caption{\textbf{Spearman correlations for SSA-COMET in MTL setting of MTE}--trained on different base models and with/without filtered training data from excluded languages. The best scores are \textbf{bolded}.}
\label{tab:114vs76}
\end{table}

\section{Conclusion}

In this work, we present \textsc{SSA-MTE}, a high-quality dataset for MT evaluation in Sub-Saharan African languages, covering 14 language pairs and over 73{,}000 human annotations. Built on this dataset, we introduce \textsc{SSA-COMET} and \textsc{SSA-COMET-QE} for MTE and QE tasks tailored to the low-resource African languages.
In our evaluation, \textsc{SSA-COMET-MTL} achieves the highest average correlation with human judgments in MTE, surpassing all prior regression-based metrics and performing competitively with the strong LLM baseline, \textit{Gemini-2.5 Pro}.

To our best knowledge, we are among the first to show that LLM prompting with just five demonstrations can yield strong evaluation performance for under-resourced languages, offering a simple and effective solution. However, it is not efficient. 
\textsc{SSA-COMET} offers a compelling solution for both MTE and QE senarios, achieving significantly higher efficiency by several orders of magnitude in inference cost (e.g., time and computational resources), while maintaining strong effectiveness when the African language is supported by the pretrained encoder.
All data, models, and code are released under open licenses (CC BY 4.0) to facilitate future research and encourage the development of inclusive, regionally adapted, and reliable evaluation tools for African languages.

\section*{Acknowledgment}
This research was supported by the Google grant via Mila. Additionally, the Portuguese to Mozambican language dataset was created with support from the Lacuna Fund and Google.org. David Adelani acknowledges the funding of IVADO and the Canada First Research Excellence Fund. We would like to extend our sincere gratitude to Markus Freitag and Parker Riley for their valuable pre-review and insightful suggestions on this paper. We would also like to thank Google Cloud for the GCP credits Award through the Gemma 2 Academic Program for LLM inference, and OpenAI for providing us access for providing API
credits through their Researcher Access API Program. Also, we thank Google Cloud for providing access to TPU v4-8 for training the AfroXLMR-114L model. We are grateful to Marek Masiak and Hao Yu for their assistance in setting up the annotation tool. Finally, we are grateful to Masakhane for their administrative support throughout the project.


\section*{Limitations}
While our work has made significant progress in MT evaluation for African languages, several limitations remain.


Moreover, our current evaluation primarily focuses on the adequacy dimension of translation quality. Future work could extend this framework to include complementary aspects such as fluency, grammaticality, terminology consistency, and discourse-level coherence, as these factors are especially important in high-stakes or professional translation scenarios.

It is worth noting that, in contrast to the findings of \citet{wang-etal-2024-afrimte}, this work reveal a relatively small performance gap between reference-based MTE models and reference-free QE models (see Tables~\ref{tab:mte_spearman_3dp} and~\ref{tab:qe-spearman-pearson}). This observation prompts a research question: \textit{as pretrained language models continue to improve in multilingual capabilities, to what extent is the presence of a reference still necessary for reliable translation evaluation?} We leave this investigation for future work.

\section*{Ethical Considerations} We employed paid annotators for this project, and paid them appropriate renumeration for their work. We pay each annotator who contributed 3,300 annotations around $\$590$, while a single translator earned $\$700$ for the translation of 1,500 sentences. When two translators are available, they earn half of the amount. We do not have other ethical issues with the source of the texts used for translation and annotation, and do not foresee any privacy issues since the source texts are from the general domain---\textit{news domain}. 

For the paper writing, ChatGPT is used only for grammar and typo errors check. 

\bibliography{custom}

\appendix


\begin{table*}[htbp]
\centering
\large
\setlength\tabcolsep{4pt}
\resizebox{1.0\textwidth}{!}{%
\begin{tabular}{c|cc|ccccc|c|c|cc}
\toprule
\makecell[l]{LP} & \textbf{Bleu} & \textbf{ChrF++} & \textbf{COMET22} & 
\makecell[c]{\textbf{AfriCOMET} \\ \textbf{v1.1 STL}} & 
\makecell[c]{\textbf{AfriCOMET} \\ \textbf{v1.0 MTL}} & 
\makecell[c]{\textbf{AfriCOMET} \\ \textbf{v1.1 MTL}} & 
\textbf{Metric-X 24} & \textbf{Claude-3.7} & \textbf{Gemini-2.5 Pro} & 
\makecell[c]{\textbf{SSA-COMET} \\ \textbf{STL}} & 
\makecell[c]{\textbf{SSA-COMET} \\ \textbf{MTL}} \\
\midrule
\texttt{eng-amh} & 0.311 & 0.446 & 0.550 & 0.622 & 0.645 & 0.651 & \textbf{0.671} & 0.602 & 0.636 & 0.602 & 0.648 \\
\texttt{eng-hau} & 0.322 & 0.421 & 0.407 & 0.474 & 0.482 & 0.481 & 0.506 & 0.445 & 0.474 & 0.469 & \textbf{0.516} \\
\texttt{eng-kik} & 0.454 & 0.586 & 0.259 & 0.495 & 0.529 & 0.688 & 0.597 & 0.638 & 0.685 & 0.689 & \textbf{0.767} \\
\texttt{eng-kin} & 0.347 & 0.500 & 0.360 & 0.585 & 0.701 & 0.662 & 0.752 & 0.698 & 0.707 & 0.750 & \textbf{0.795} \\
\texttt{eng-luo} & 0.408 & 0.590 & 0.368 & 0.604 & 0.501 & 0.685 & 0.535 & 0.648 & 0.758 & 0.660 & \textbf{0.770} \\
\texttt{eng-twi} & 0.283 & 0.496 & 0.444 & 0.628 & 0.634 & 0.698 & 0.723 & 0.748 & \textbf{0.779} & 0.718 & 0.774 \\
\texttt{eng-yor} & 0.331 & 0.455 & 0.378 & 0.497 & 0.583 & 0.540 & 0.572 & 0.591 & 0.600 & 0.610 & \textbf{0.668} \\
\texttt{fra-ewe} & 0.201 & 0.346 & 0.307 & 0.418 & 0.479 & 0.514 & 0.511 & 0.544 & 0.569 & 0.535 & \textbf{0.599} \\
\texttt{fra-wol} & 0.399 & 0.568 & 0.331 & 0.456 & 0.475 & 0.474 & 0.541 & 0.649 & \textbf{0.707} & 0.618 & 0.703 \\
\texttt{por-vmw} & 0.160 & 0.437 & 0.253 & 0.305 & 0.350 & 0.273 & 0.478 & 0.526 & \textbf{0.535} & 0.467 & 0.527 \\
\texttt{por-nya} & 0.460 & 0.827 & 0.803 & 0.912 & 0.915 & 0.904 & 0.931 & 0.887 & 0.892 & 0.920 & \textbf{0.932} \\
\midrule
Average          & 0.334 & 0.516 & 0.405 & 0.545 & 0.572 & 0.597 & 0.620 & 0.634 & 0.667 & 0.640 & \textbf{0.700} \\
\bottomrule
\end{tabular}
}
\caption{Pearson correlation of MTE metrics across language pairs. The best scores are \textbf{bolded}.}
\label{tab:mte_pearson_3dp}
\end{table*}

\begin{table}[htbp]
\centering
\scriptsize
\begin{tabular}{l|cc|cc}
\toprule
 
\multirow{2}{*}{\textbf{LP}}
& \multicolumn{2}{c|}{\textbf{SSA-COMET-STL}} 
& \multicolumn{2}{c}{\textbf{SSA-COMET-MTL}} \\
\cmidrule(lr){2-3} \cmidrule(lr){4-5}
& \textbf{w/ WMT}& \textbf{w/o WMT}   
& \textbf{w/ WMT}& \textbf{w/o WMT}   \\
\midrule
\texttt{eng-amh}  & 0.602 & 0.575 & \textbf{0.648} & 0.573 \\
\texttt{eng-hau}  & 0.469 & 0.448 & \textbf{0.516} & 0.447 \\
\texttt{eng-kik}  & 0.689 & 0.697 & \textbf{0.767} & 0.748 \\
\texttt{eng-kin}  & \textbf{0.750} & 0.744 & 0.795 & 0.783 \\
\texttt{eng-luo}  & 0.660 & 0.652 & \textbf{0.770} & 0.743 \\
\texttt{eng-twi}  & 0.718 & 0.713 & \textbf{0.774} & 0.742 \\
\texttt{eng-yor}  & 0.610 & 0.602 & \textbf{0.668} & 0.625 \\
\texttt{fra-ewe}  & 0.535 & 0.549 & \textbf{0.599} & 0.566 \\
\texttt{fra-wol}  & 0.618 & 0.636 & \textbf{0.703} & 0.693 \\
\texttt{por-vmw}  & \textbf{0.467} & 0.436 & 0.527 & 0.480 \\
\texttt{por-nya}  & 0.920 & 0.893 & \textbf{0.932} & 0.915 \\
\midrule
\textbf{Average} 
         & 0.640 & 0.631 & \textbf{0.700} & 0.665 \\
\bottomrule
\end{tabular}
\caption{Pearson correlations for SSA-COMET-STL and SSA-COMET-MTL trained with and without WMT data. The best scores are \textbf{bolded}.}
\label{tab:pearson-wmt}
\end{table}

\begin{figure}[htbp]
  \centering
  \includegraphics[
    width=1.5\linewidth,
    trim=11cm 0cm 0cm 0cm, 
    clip
  ]{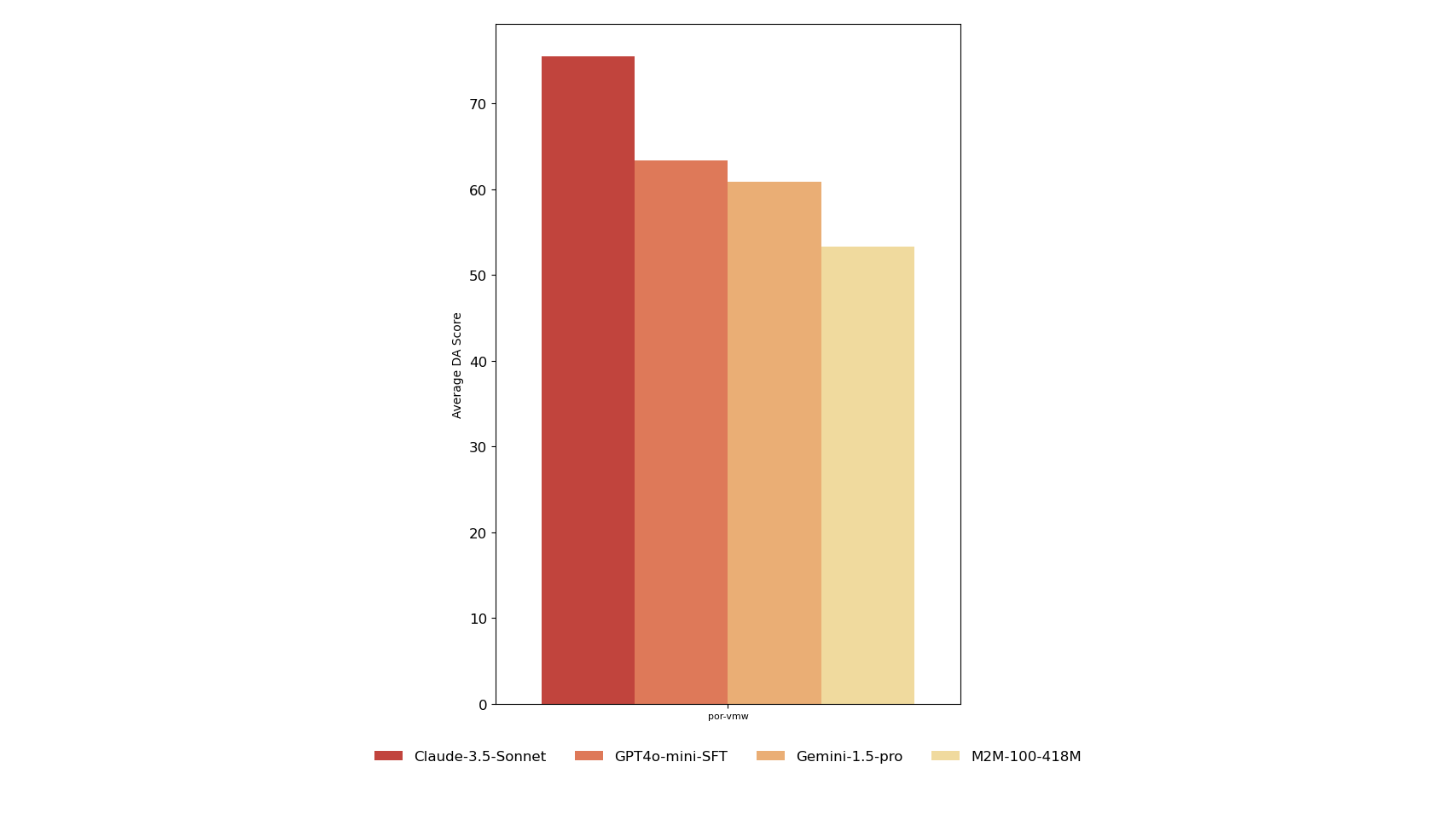}
  \caption{Translation performance of MT systems used for \texttt{por-vmw}.}
  \label{fig:systems_vmw}
\end{figure}

\begin{figure}[htbp]
  \centering
  \includegraphics[
    width=1.5\linewidth,
    trim=11cm 0cm 0cm 0cm, 
    clip
  ]{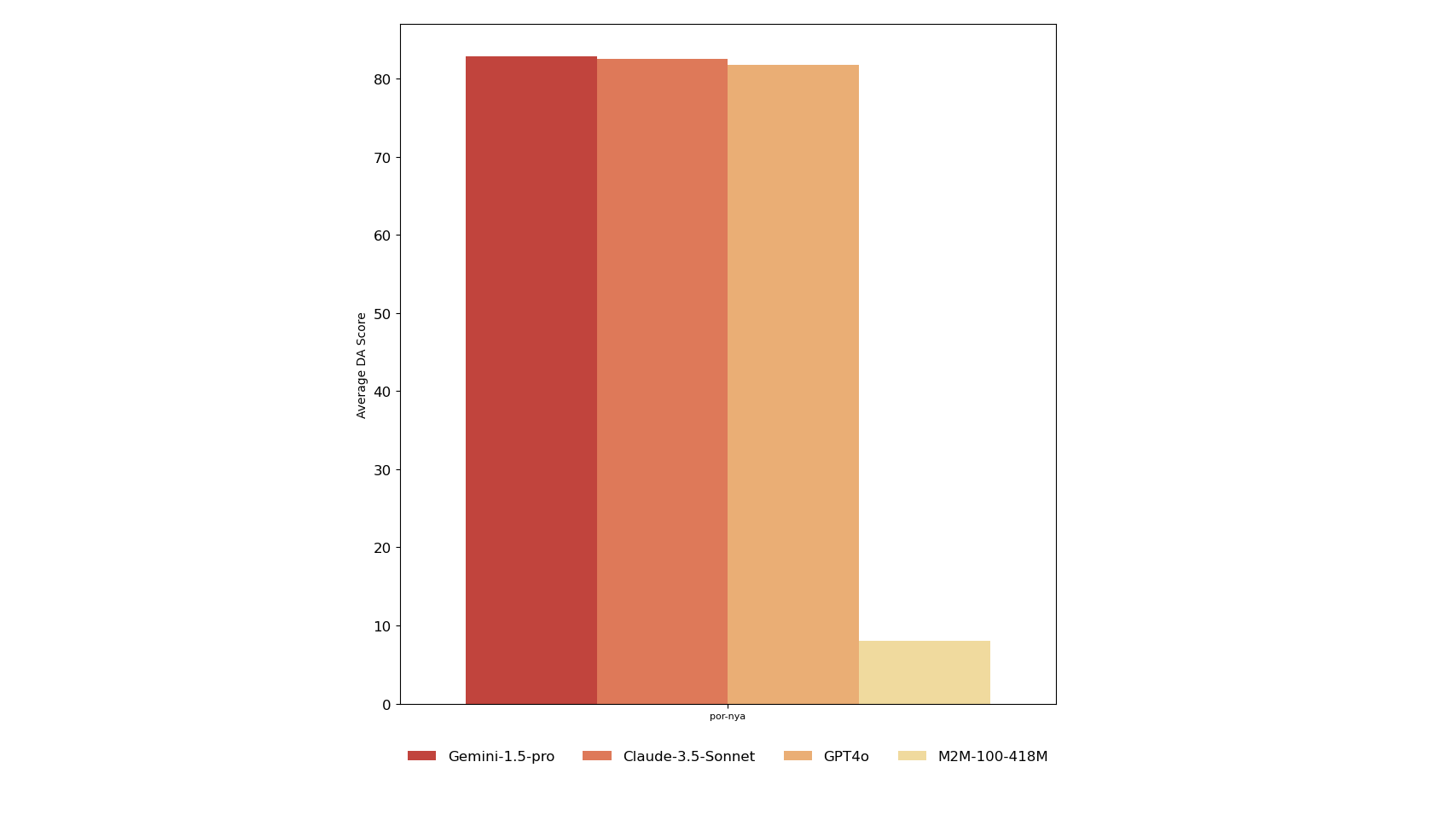}
  \caption{Translation performance of MT systems used for \texttt{por-nya}.}
  \label{fig:systems_nya}
\end{figure}

\section{Correlations between number of errors and the final scores}
Table~\ref{tab:zscore-corr} presents the correlation between Z-normalized DA scores and the frequency of different error types. Among all error categories, mistranslation shows the strongest negative correlation with overall adequacy (Spearman: $–0.521$), followed by addition and omission errors. The aggregated total error count exhibits the highest overall correlation (Spearman: –$0.574$), confirming that as the number of annotated errors increases, the adequacy score consistently decreases. These findings validate the reliability of error span annotations as strong indicators of perceived translation quality.

\begin{table}[ht]
\centering
\small
\begin{tabular}{l|cc}
\toprule
\multirow{2}{*}{\textbf{Criterion}} & \multicolumn{2}{c}{\textbf{Z-score}} \\
\cmidrule(lr){2-3}
 & \textbf{Spearman} & \textbf{Kendall} \\
\midrule
Mistranslation  &  -0.521 &  -0.377 \\
Omission        & -0.265 &  -0.210 \\
Addition       & -0.276 &   -0.218 \\
Untranslated  & -0.048 &  -0.038 \\
\midrule
\textbf{Total Error}  & \textbf{-0.574} & \textbf{-0.406} \\
\bottomrule
\end{tabular}
\caption{Correlation of each error criterion with Z-scores.}
\label{tab:zscore-corr}
\end{table}

\section{Results on AfriMTE}
To evaluate the generalization capability of our SSA-COMET models beyond the newly collected SSA-MTE dataset, we conduct experiments on the AfriMTE benchmark~\citep{wang-etal-2024-afrimte}. As shown in Table~\ref{tab:africomet_ssa_spearman_compact_spearman} and Table~\ref{tab:africomet_ssa}, \textsc{SSA-COMET-MTL} outperforms all previous AfriCOMET variants, including the strongest one, \textsc{AfriCOMET-v1.1-MTL}. 
These results demonstrate that SSA-COMET models remain robust and effective under domain shift. 
\begin{table}[htbp]
\centering
\scriptsize
\setlength\tabcolsep{3pt}
\begin{tabular}{lcccc}
\toprule
\textbf{LP} 
& \makecell[c]{\textbf{AfriCOMET} \\ \textbf{v1.1 STL}}
& \makecell[c]{\textbf{AfriCOMET} \\ \textbf{v1.1 MTL}}
& \makecell[c]{\textbf{SSA-COMET} \\ \textbf{STL}}
& \makecell[c]{\textbf{SSA-COMET} \\ \textbf{MTL}} \\
\midrule
\texttt{ary-fra}  & 0.526 & 0.561 & 0.505 & \textbf{0.586} \\
\texttt{eng-arz}  & 0.510 & 0.579 & 0.525 & \textbf{0.600} \\
\texttt{eng-fra}  & 0.492 & 0.507 & 0.505 & \textbf{0.553} \\
\texttt{eng-hau}  & 0.561 & \textbf{0.614} & 0.536 & 0.596 \\
\texttt{eng-ibo}  & 0.522 & \textbf{0.582} & 0.529 & 0.491 \\
\texttt{eng-kik}  & 0.430 & 0.520 & 0.445 & \textbf{0.590} \\
\texttt{eng-luo}  & 0.325 & \textbf{0.515} & 0.412 & 0.502 \\
\texttt{eng-som}  & 0.502 & 0.525 & 0.519 & \textbf{0.529} \\
\texttt{eng-swh}  & 0.704 & 0.756 & 0.708 & \textbf{0.761} \\
\texttt{eng-twi}  & 0.222 & 0.209 & 0.193 & \textbf{0.223} \\
\texttt{eng-xho}  & 0.203 & 0.157 & \textbf{0.205} & 0.137 \\
\texttt{eng-yor}  & 0.338 & 0.473 & 0.394 & \textbf{0.548} \\
\texttt{yor-eng}  & 0.508 & \textbf{0.566} & 0.451 & 0.548 \\
\midrule
\textbf{Average}  & 0.449 & 0.505 & 0.456 & \textbf{0.513} \\
\bottomrule
\end{tabular}
\caption{Spearman correlation of AfriCOMET and SSA-COMET on AfriMTE. The best scores are \textbf{bolded}.}
\label{tab:africomet_ssa_spearman_compact_spearman}
\end{table}

\begin{table}[htbp]
\centering
\scriptsize
\setlength\tabcolsep{3pt}
\begin{tabular}{lcccc}
\toprule
\textbf{LP} 
& \makecell[c]{\textbf{AfriCOMET} \\ \textbf{v1.1 STL}}
& \makecell[c]{\textbf{AfriCOMET} \\ \textbf{v1.1 MTL}}
& \makecell[c]{\textbf{SSA-COMET} \\ \textbf{STL}}
& \makecell[c]{\textbf{SSA-COMET} \\ \textbf{MTL}} \\
\midrule
\texttt{ary-fra}  & 0.553 & 0.641 & 0.547 & \textbf{0.661} \\
\texttt{eng-arz}  & 0.515 & \textbf{0.603} & 0.515 & 0.600 \\
\texttt{eng-fra}  & 0.544 & 0.484 & 0.517 & \textbf{0.526} \\
\texttt{eng-hau}  & \textbf{0.647} & 0.613 & 0.606 & 0.608 \\
\texttt{eng-ibo}  & 0.496 & \textbf{0.664} & 0.558 & 0.612 \\
\texttt{eng-kik}  & 0.686 & 0.545 & 0.689 & \textbf{0.696} \\
\texttt{eng-luo}  & 0.480 & 0.526 & 0.543 & \textbf{0.621} \\
\texttt{eng-som}  & 0.460 & 0.374 & \textbf{0.465} & 0.410 \\
\texttt{eng-swh}  & 0.737 & 0.762 & 0.740 & \textbf{0.817} \\
\texttt{eng-twi}  & 0.474 & 0.296 & 0.410 & \textbf{0.486} \\
\texttt{eng-xho}  & 0.384 & 0.345 & 0.413 & \textbf{0.507} \\
\texttt{eng-yor}  & 0.595 & 0.634 & 0.616 & \textbf{0.694} \\
\texttt{yor-eng}  & 0.521 & \textbf{0.571} & 0.455 & \textbf{0.571} \\
\midrule
\textbf{Average}  & 0.545 & 0.543 & 0.544 & \textbf{0.601} \\
\bottomrule
\end{tabular}
\caption{Pearson correlation of AfriCOMET-V1.1 and SSA-COMET on AfriMTE. The best scores are \textbf{bolded}.}
\label{tab:africomet_ssa}
\end{table}

\begin{table*}[ht]
\centering
\small
\begin{tabular}{l|cc|cc|cc}
\toprule
\textbf{Metric} & \makecell[c]{Gemini-2.0 Flash\\0-shot} & \makecell[c]{Gemini-2.0 Flash\\5-shot} & \makecell[c]{LLaMA4 400B\\0-shot} & \makecell[c]{LLaMA4 400B\\5-shot} & \makecell[c]{Claude-3.7\\0-shot} & \makecell[c]{Claude-3.7\\5-shot} \\
\midrule
Spearman & 0.468 & 0.544 & 0.325 & 0.513 & 0.470 & 0.583 \\

Pearson  & 0.506 & 0.575 & 0.368 & 0.551 & 0.499 & 0.609 \\
\bottomrule
\end{tabular}
\caption{Performance differences of LLMs in Zero-shot vs. 5-shot prompting on SSA-MTE. (Results were obtained without Nyanja)}
\label{tab:llm_fewshot_comparison}
\end{table*}


\begin{table}[ht]
\centering
\small
\begin{tabular}{l|cc|cc}
\toprule
\multirow{2}{*}{\textbf{LP}} & \multicolumn{2}{c|}{\textbf{Evaluator 1}} & \multicolumn{2}{c}{\textbf{Evaluator 2}} \\
\cmidrule(lr){2-3} \cmidrule(lr){4-5}
 & \textbf{Spear.} & \textbf{Pear.} & \textbf{Spear.} & \textbf{Pear.} \\
\midrule
\texttt{eng-ibo} & 0.392 & 0.447 & 0.277 & 0.357 \\
\texttt{eng-zul} & 0.321 & 0.363 & 0.273 & 0.341 \\
\bottomrule
\end{tabular}
\caption{Per-annotator Spearman-rank and Pearson correlations with silver references produced by AfriCOMET trained with 8 LPs.}
\label{tab:annotator_correlation}
\end{table}
\begin{table*}[htbp]
\centering
\large
\setlength\tabcolsep{4pt}
\resizebox{1\textwidth}{!}{%
\begin{tabular}{c|cccc|cccc}
\toprule
\makecell[l]{LP} 
& \makecell[c]{\textbf{Gemma3-27B-it}} 
& \makecell[c]{\textbf{Llama4 100B}} 
& \makecell[c]{\textbf{Llama4 400B}} 
& \makecell[c]{\textbf{Deepseek V3}} 
& \makecell[c]{\textbf{GPT4o}} 
& \makecell[c]{\textbf{Gemini-2.0 Flash}} 
& \makecell[c]{\textbf{Claude-3.7}} 
& \makecell[c]{\textbf{Gemini-2.5 Pro}} \\
\midrule
\texttt{eng-amh} & 0.503 & 0.391 & 0.505 & 0.496 & 0.429 & 0.513 & 0.566 & 0.605 \\
\texttt{eng-hau} & 0.337 & 0.323 & 0.333 & 0.373 & 0.411 & 0.299 & 0.425 & 0.471 \\
\texttt{eng-kik} & 0.573 & 0.566 & 0.666 & 0.639 & 0.635 & 0.687 & 0.696 & 0.735 \\
\texttt{eng-kin} & 0.467 & 0.426 & 0.455 & 0.461 & 0.498 & 0.492 & 0.536 & 0.528 \\
\texttt{eng-luo} & 0.489 & 0.524 & 0.610 & 0.549 & 0.639 & 0.699 & 0.678 & 0.782 \\
\texttt{eng-twi} & 0.504 & 0.532 & 0.578 & 0.560 & 0.571 & 0.613 & 0.652 & 0.710 \\
\texttt{eng-yor} & 0.399 & 0.422 & 0.435 & 0.434 & 0.421 & 0.375 & 0.501 & 0.524 \\
\texttt{fra-ewe} & 0.430 & 0.434 & 0.539 & 0.546 & 0.461 & 0.621 & 0.614 & 0.658 \\
\texttt{fra-wol} & 0.509 & 0.519 & 0.645 & 0.647 & 0.663 & 0.712 & 0.699 & 0.750 \\
\texttt{por-vmw} & 0.315 & 0.322 & 0.363 & 0.275 & 0.330 & 0.431 & 0.463 & 0.487 \\
\texttt{por-nya} & 0.648 & 0.635 & 0.650 & 0.672 & 0.647 & 0.678 & 0.684 & 0.709 \\
\midrule
Average         & 0.470 & 0.463 & 0.525 & 0.514 & 0.519 & 0.556 & 0.592 & 0.633 \\
\bottomrule
\end{tabular}
}
\caption{Spearman correlation of different LLM-based metrics across LPs without generating error spans. }
\label{tab:llm_spearman_final}
\end{table*}

\begin{table*}[htbp]
\centering
\large
\setlength\tabcolsep{4pt}
\resizebox{1\textwidth}{!}{%
\begin{tabular}{c|cccccccc}
\toprule
\makecell[l]{LP} 
& \makecell[c]{\textbf{Gemma3-27B-it}} 
& \makecell[c]{\textbf{Llama4 100B}} 
& \makecell[c]{\textbf{Llama4 400B}} 
& \makecell[c]{\textbf{Deepseek V3}} 
& \makecell[c]{\textbf{GPT4o}} 
& \makecell[c]{\textbf{Gemini-2.0 Flash}} 
& \makecell[c]{\textbf{Claude-3.7}} 
& \makecell[c]{\textbf{Gemini-2.5 Pro}} \\
\midrule
\texttt{eng-amh} & 0.556 & 0.456 & 0.551 & 0.551 & 0.488 & 0.576 & 0.602 & 0.636 \\
\texttt{eng-hau} & 0.348 & 0.350 & 0.398 & 0.390 & 0.421 & 0.338 & 0.445 & 0.474 \\
\texttt{eng-kik} & 0.518 & 0.491 & 0.608 & 0.530 & 0.551 & 0.607 & 0.638 & 0.685 \\
\texttt{eng-kin} & 0.648 & 0.594 & 0.662 & 0.649 & 0.675 & 0.679 & 0.698 & 0.707 \\
\texttt{eng-luo} & 0.473 & 0.498 & 0.584 & 0.517 & 0.595 & 0.650 & 0.648 & 0.758 \\
\texttt{eng-twi} & 0.638 & 0.664 & 0.685 & 0.688 & 0.702 & 0.733 & 0.748 & 0.779 \\
\texttt{eng-yor} & 0.553 & 0.552 & 0.571 & 0.576 & 0.569 & 0.509 & 0.591 & 0.600 \\
\texttt{fra-ewe} & 0.382 & 0.386 & 0.473 & 0.462 & 0.421 & 0.523 & 0.544 & 0.569 \\
\texttt{fra-wol} & 0.438 & 0.432 & 0.555 & 0.542 & 0.591 & 0.624 & 0.649 & 0.707 \\
\texttt{por-vmw} & 0.424 & 0.424 & 0.423 & 0.398 & 0.429 & 0.512 & 0.526 & 0.535 \\
\texttt{por-nya} & 0.907 & 0.861 & 0.879 & 0.854 & 0.878 & 0.910 & 0.887 & 0.892 \\
\midrule
Average         & 0.535 & 0.519 & 0.581 & 0.560 & 0.575 & 0.606 & 0.634 & 0.667 \\
\bottomrule
\end{tabular}
}
\caption{Pearson correlation of different LLM-based metrics across LPs without generating error spans.}
\label{tab:llm_pearson_corrected}
\end{table*}

\begin{table*}[htbp]
\centering
\scriptsize
\setlength\tabcolsep{4pt}
\begin{tabular}{l|cccccccc}
\toprule
\textbf{LP} 
& \textbf{Gemma3-27B-it} 
& \textbf{Llama4 100B} 
& \textbf{Llama4 400B} 
& \textbf{Deepseek V3} 
& \textbf{GPT-4o} 
& \textbf{Gemini-2.0 Flash} 
& \textbf{Claude-3.7} 
& \textbf{Gemini-2.5 Pro} \\
\midrule
\texttt{eng-amh} & 0.359 & 0.261 & 0.486 & --    & 0.335 & 0.472 & 0.527 & 0.598 \\
\texttt{eng-hau} & 0.267 & 0.152 & 0.310 & 0.167 & 0.271 & 0.289 & 0.407 & 0.450 \\
\texttt{eng-kik} & 0.422 & 0.425 & 0.645 & 0.539 & 0.455 & 0.691 & 0.702 & 0.707 \\
\texttt{eng-kin} & 0.497 & 0.281 & 0.514 & 0.339 & 0.465 & 0.540 & 0.500 & 0.529 \\
\texttt{eng-luo} & 0.240 & 0.233 & 0.538 & 0.260 & 0.357 & 0.615 & 0.739 & 0.712 \\
\texttt{eng-twi} & 0.384 & 0.353 & 0.542 & 0.415 & 0.415 & --    & 0.662 & 0.649 \\
\texttt{eng-yor} & 0.373 & 0.237 & 0.452 & 0.396 & 0.343 & 0.450 & 0.470 & 0.522 \\
\texttt{fra-ewe} & 0.334 & 0.210 & 0.489 & 0.307 & 0.221 & 0.590 & 0.581 & 0.631 \\
\texttt{fra-wol} & 0.341 & 0.347 & 0.578 & 0.463 & 0.422 & 0.671 & 0.704 & 0.724 \\
\texttt{por-vmw} & 0.205 & 0.158 & 0.284 & 0.101 & 0.066 & 0.256 & 0.474 & 0.383 \\
\texttt{por-nya} & 0.595 & 0.614 & 0.651 & 0.549 & 0.619 & 0.637 & 0.664 & 0.645 \\
\midrule
\textbf{Average} & 0.365 & 0.298 & 0.499 & 0.354 & 0.361 & 0.521 & 0.585 & 0.595 \\
\bottomrule
\end{tabular}
\caption{Spearman correlation of LLM-based metrics across language pairs, using error span prediction. ``-'' indicates that the model's output failed or collapsed for that LP.}
\label{tab:llm_spearman_3dp}
\end{table*}

\begin{table*}[htbp]
\centering
\scriptsize
\setlength\tabcolsep{4pt}
\begin{tabular}{l|cccccccc}
\toprule
\textbf{LP} 
& \textbf{Gemma3-27B-it} 
& \textbf{Llama4 100B} 
& \textbf{Llama4 400B} 
& \textbf{Deepseek V3} 
& \textbf{GPT-4o} 
& \textbf{Gemini-2.0 Flash} 
& \textbf{Claude-3.7} 
& \textbf{Gemini-2.5 Pro} \\
\midrule
\texttt{eng-amh} & 0.382 & 0.296 & 0.503 & --    & 0.333 & 0.514 & 0.568 & 0.611 \\
\texttt{eng-hau} & 0.262 & 0.145 & 0.353 & 0.157 & 0.280 & 0.322 & 0.434 & 0.456 \\
\texttt{eng-kik} & 0.400 & 0.404 & 0.599 & 0.472 & 0.427 & 0.629 & 0.651 & 0.706 \\
\texttt{eng-kin} & 0.584 & 0.283 & 0.607 & 0.356 & 0.529 & 0.628 & 0.642 & 0.692 \\
\texttt{eng-luo} & 0.258 & 0.243 & 0.531 & 0.278 & 0.363 & 0.591 & 0.721 & 0.713 \\
\texttt{eng-twi} & 0.459 & 0.357 & 0.609 & 0.448 & 0.473 & --    & 0.741 & 0.695 \\
\texttt{eng-yor} & 0.493 & 0.293 & 0.523 & 0.488 & 0.420 & 0.545 & 0.594 & 0.588 \\
\texttt{fra-ewe} & 0.303 & 0.160 & 0.459 & 0.271 & 0.233 & 0.545 & 0.537 & 0.592 \\
\texttt{fra-wol} & 0.335 & 0.346 & 0.527 & 0.438 & 0.407 & 0.625 & 0.652 & 0.702 \\
\texttt{por-vmw} & 0.252 & 0.161 & 0.336 & 0.155 & 0.141 & 0.292 & 0.527 & 0.441 \\
\texttt{por-nya} & 0.839 & 0.851 & 0.892 & 0.758 & 0.813 & 0.856 & 0.884 & 0.889 \\
\midrule
\textbf{Average} & 0.415 & 0.322 & 0.540 & 0.382 & 0.402 & 0.555 & 0.632 & 0.644 \\
\bottomrule
\end{tabular}
\caption{Pearson correlation of LLM-based metrics across language pairs, using error span prediction. ``-'' indicates that the model's output failed or collapsed for that LP.}
\label{tab:llm_pearson_3dp}
\end{table*}

\section{Data Collection Process for the Portuguese Texts}
\label{sec:appendix-por-source}

The Portuguese sentences were sourced from the Multilingual Open Text dataset \cite{palen-michel-etal-2022-multilingual}, which features news articles published by Voice of America (VOA\footnote{\url{https://www.voanews.com/}}). These sentences were translated into Emakhuwa, resulting in a parallel corpus that was released under a CC BY 4.0 license and made publicly available in \citet{ali-etal-2024-building}. The dataset has three splits, TRAIN, DEV, and TEST, and covers seven topics: politics, economy, culture, sports, health, society, and world news. We only focus on the annotations for the Test split in this study due to constraints of annotation resources.

\section{Machine Translations for Emakhuwa}
\label{sec:appendix-MT-vma}
We sampled 1,128 parallel sentences from the Test split of the Portuguese–Emakhuwa dataset. The source sentences were used to generate translations from Portuguese into Emakhuwa using the machine translation systems in Figure~\ref{fig:systems_vmw}. 

\section{Emakhuwa Data annotation process}
\label{sec:appendix-process-vma}
For the more challenging Portuguese-source language pairs, \texttt{por-vmw}, we annotated 1,600 samples evenly distributed between 2 evaluators, with 300 overlapping samples split between two evaluators for quality control. 

\section{Further selection of Training Data for Zulu and Igbo}

\label{appedix-further-selection}
We hypothesize that the low agreement may be due to one evaluator consistently outperforming the other in annotation quality. To address this, we retained only the annotations from the more reliable evaluator for inclusion in the training set. Building on the success of AfriCOMET~\citep{wang-etal-2024-afrimte}, we employed a COMET model trained with AfroXLMR-76L encoder backbone, using a multi-task learning framework~\citep{wang-etal-2024-afrimte, wan2022unite}, on eight language pairs that achieved both Spearman and ICC scores above 0.48 in Table~\ref{tab:correlation_icc}: eng-amh, eng-kik, eng-kin, eng-luo, eng-twi, eng-yor, fra-ewe, and fra-wol. We then used this model to generate predicted scores for \texttt{eng-ibo}, \texttt{eng-swa}, and \texttt{eng-zul}, combined with WMT DA data. The resulting model served as a silver labeller, which the generated scores were used as a silver reference for evaluating annotator reliability. Next, we compared the model-generated scores with those from each evaluator individually and computed both Spearman rank and Pearson correlation coefficients. The results, presented in Table~\ref{tab:annotator_correlation}, reveal clear gaps in correlation: evaluator 1 for Zulu and Igbo, and evaluator 2 for Swahili, consistently show higher agreement with the silver reference. Therefore, we include their annotations for \texttt{eng-ibo} and \texttt{eng-zul} in the training set of SSA-MTE. 


\section{Handling for unexpected outputs from LLMs}
\label{appendix-LLM-handling}
For a small number of cases, the LLMs fail to generate a valid answer and instead return an uninterpretable response. Since our evaluation operates within a normalized range of 
[0,1], we assign a default score of $0.5$—representing a neutral judgment—for these cases. This approach ensures that failing cases do not disproportionately affect overall results, while preserving the integrity of the evaluation. Discarding such cases could introduce selection bias, obscure model weaknesses, and compromise comparability across systems.

\section{More Details on the Prompting}
\label{sec:zero-shot-prompting}

For all prompting experiments, we used the default decoding settings provided by the API of each LLM. We did not enforce greedy decoding or adjust temperature, top-$p$, or other sampling parameters. This ensures the results reflect realistic usage scenarios, where users rely on default behavior without fine-tuning generation strategies.

For the 0-shot prompting setup, we removed all demonstration-related content from the prompt, leaving only the annotation guideline and the final instruction for predicting the adequacy score.

Table \ref{tab:llm_fewshot_comparison} compares zero-shot and few-shot results, the results shows that without demonstration examples, the performance of the LLMs are unreliable, and far below the performance of SSA-COMET models.

 \section{Comparison: Gemini-2.5 Pro vs. SSA-COMET-MTL}

Under the MTE setting, Gemini-2.5 Pro and SSA-COMET-MTL achieve similar overall Spearman correlation. However, when excluding the \texttt{por-vmw} language pair, which is not covered in the pretraining data of the encoder used in SSA-COMET—SSA-COMET-MTL demonstrates a clear advantage, with an average Spearman score that is $0.019$ higher. This margin of improvement is comparable to the performance gap between Gemini-2.5 Pro with and without reference input.

Moreover, even when including \texttt{por-vmw}, SSA-COMET-MTL clearly outperforms Gemini-2.5 Pro in terms of Pearson correlation, with a margin of $0.023$. This indicates that SSA-COMET-MTL produces adequacy scores that are more accurately aligned with human ratings in absolute terms, not just in relative ranking.

Under the QE setting, excluding the \texttt{por-vmw} language pair from the average, SSA-COMET-MTL achieves a slightly higher Spearman correlation and a notably stronger Pearson correlation, with an advantage of $0.0302$. 

These results suggest that for languages covered by the encoder, SSA-COMET-MTL is not only more accurate but also significantly more efficient than Gemini-2.5 Pro. On the SSA-MTE test set, SSA-COMET evaluates each LP in under two minutes, whereas prompting LLMs requires substantially more time per sample. This makes SSA-COMET a more scalable and practical solution for low-resource MT evaluation.

\section{Qualitative evaluation of LLM Error-Span Predictions}
\label{sec:qualitative}
Table ~\ref{tab:qualitative} shows three examples of the predictions of Gemini-2.5 Pro and Llama 4 400 B. We find that the former aligns more with the human judgements than the latter, which aligns with our prompting results in Table ~\ref{tab:error-span-results}. Furthermore, we find the error span predictions to be helpful in many cases. We leave a more detailed investigation for future work.

\begin{table*}[ht]
\centering
\footnotesize
\resizebox{\textwidth}{!}{
\begin{tabular}{p{150mm}|r}
\toprule
\textbf{Sentence / LLM output} & \textbf{Human rating} \\
\midrule
\textbf{Source:} Angola, as everyone knows, is a beautiful country. &  \\
\rowcolor{LightCyan}
\textbf{Translation:} G\d{e}g\d{e}bi gbogbo eniyan ti m\d{o}, Angola j\d{e} oril\d{e}-ede ti o \textcolor{red}{dara jul\d{o}}.  &  0.77 \\
\textbf{Translation (in English)} As everyone knows, Angola is \textcolor{red}{the most} beautiful country. &  \\

\textbf{Reference:} Angola, b\'{i} gbogbo \`{e}n\`{i}y\`{a}n \d{s}e m\d{\`o}, j\d{\'e} or\'{i}l\d{\`e}-\`{e}d\`{e} tí \'{o} r\d{e}w\`{a}.  \\

\addlinespace
\textbf{Gemini 2.5 Pro Output} \\
\textbf{Output:} The following error is detected: \\
\textbf{Mistranslation:} The span "beautiful" in the source text was wrongly translated into "dara jul\d{o}" in the translation. \\
\textbf{Source:} Angola, as everyone knows, is a <s>beautiful</s> country. \\
\textbf{Target:} G\d{e}g\d{e}bi gbogbo eniyan ti m\d{o}, Angola j\d{e} oril\d{e}-ede ti o <s>dara jul\d{o}</s>. \\

Based on the 1 error detected, \textbf{the score of translation is: 0.67} \\

\addlinespace
\textbf{LLama 4 400B Output} \\
The following error is detected: \\
\textbf{Mistranslation:} The span "beautiful country" in the source text was wrongly translated into "oril\d{e}-ede ti o dara jul\d{o}" in the translation.\\
\textbf{Source:} Angola, as everyone knows, is a <s>beautiful country. </s>  \\
\textbf{Target:} G\d{e}g\d{e}bi gbogbo eniyan ti m\d{o}, Angola j\d{e} <s> oril\d{e}-ede ti o dara jul\d{o}</s>. \\

Based on the 1 error detected, \textbf{the score of translation is: 0.58} \\

\midrule


\textbf{Source:} Nobody knew Alffa outside a small part of Wales &  \\
\rowcolor{LightCyan}
\textbf{Translation:} K\`{o} s\'{i} \d{e}ni t\'{o} m\d{o} Alffa \textcolor{red}{n\'i \`{a}gb\`{e}gb\`{e} k\'{e}ker\'{e} kan} n\'{i} Wales  &  0.76 \\
\textbf{Translation (in English)} Nobody knew Alffa \textcolor{red}{in this small area} of Wales &  \\

\textbf{Reference:} K\`{o} s\'{i} \d{e}ni t\'{o} m\d{o} Alffa y\`{a}t\d{\`o} s\'{i} \`aw\d{o}n ibi k\'er\'eje n\'i Wales \\

\addlinespace
\textbf{Gemini 2.5 Pro Output} \\
\textbf{Output:} The following error is detected: \\
\textbf{Mistranslation:} The span ""outside"" in the source text was wrongly translated into ""n\'{i}"" in the translation. \\
\textbf{Source:} Nobody knew Alffa <s>outside</s> a small part of Wales\\
\textbf{Target:}  K\`{o} s\'{i} \d{e}ni t\'{o} m\d{o} Alffa <s>n\'i</s> \`{a}gb\`{e}gb\`{e} k\'{e}ker\'{e} kan n\'{i} Wales  \\

Based on the 1 error detected, \textbf{the score of translation is: 0.67} \\

\addlinespace
\textbf{LLama 4 400B Output} \\
The following error is detected: \\
\textbf{Mistranslation:} The span ""outside a small part of Wales"" in the source text was wrongly translated into ""n\'i \`{a}gb\`{e}gb\`{e} k\'{e}ker\'{e} kan n\'{i} Wales"" in the translation. \\
\textbf{Source:} Nobody knew Alffa <s>outside a small part of Wales</s>\\
\textbf{Target:}  K\`{o} s\'{i} \d{e}ni t\'{o} m\d{o} Alffa <s>n\'i \`{a}gb\`{e}gb\`{e} k\'{e}ker\'{e} kan n\'{i} Wales </s>  \\

Based on the 1 error detected, \textbf{the score of translation is: 0.38} \\

\midrule

\textbf{Source:} However, professional freelance photojournalists are still quite rare. &  \\
\rowcolor{LightCyan}
\textbf{Translation:} Sib\d{e}sib\d{e}, \textcolor{red}{aw\d{o}n onise iroyin ti o ni im\d{o}-\d{e}r\d{o} ti o ni im\d{o}-\d{e}r\d{o} ti o ni im\d{o}-im\d{o}-im\d{o}-im\d{o}-im\d{o}-im\d{o}-iim\d{o}-im\d{o}-im\d{o}-im\d{o}-im\d{o}-im\d{o}-... } &  0.11 \\
\textbf{Translation (in English)} However, \textcolor{red}{the news staffs that have technical knowledge that have technical knowledge knowledge knowledge knowledge knowledge knowledge knowledge ... } &  \\

\textbf{Reference:} \d{S}\`{u}gb\d{o}\'{n}, \`{a}w\d{o}n ak\d{\'o}\d{s}\d{e}m\d{o}\d{s}\d{\'e} af\`aw\`or\'{a}nr\`oy\`in ad\'ad\'ur\'o k\`o s\`i w\d{\'o}p\d{\`o} n\'a\`a.  \\

\addlinespace
\textbf{Gemini 2.5 Pro Output} \\
\textbf{Output:} The following error is detected: \\
\textbf{Mistranslation:} the span "professional freelance photojournalists are still quite rare" in the source text was wrongly translated into "w\d{o}n onise iroyin ti o ni im\d{o}-\d{e}r\d{o} ti o ni im\d{o}-\d{e}r\d{o} ti o ni im\d{o}-im\d{o}-im\d{o}-im\d{o}-im\d{o}-im\d{o}-iim\d{o}-im\d{o}-im\d{o}-im\d{o}-im\d{o}-im\d{o}-..." in the translation. \\
\textbf{Source:} However, <s>professional freelance photojournalists are still quite rare</s> \\
\textbf{Target:} Sib\d{e}sib\d{e} <s>aw\d{o}n onise iroyin ti o ni im\d{o}-\d{e}r\d{o} ti o ni im\d{o}-\d{e}r\d{o} ti o ni im\d{o}-im\d{o}-im\d{o}-im\d{o}-im\d{o}-im\d{o}-iim\d{o}-im\d{o}-im\d{o}-im\d{o}-im\d{o}-im\d{o}-...</s> \\

Based on the 1 error detected, \textbf{the score of translation is: 0.16} \\

\addlinespace
\textbf{LLama 4 400B Output} \\
The following error is detected: \\
\textbf{Mistranslation:} The span "professional freelance photojournalists" in the source text was wrongly translated into "w\d{o}n onise iroyin ti o ni im\d{o}-\d{e}r\d{o} ti o ni im\d{o}-\d{e}r\d{o} ti o ni im\d{o}-im\d{o}-im\d{o}-im\d{o}-im\d{o}-im\d{o}-iim\d{o}-im\d{o}-im\d{o}-im\d{o}-im\d{o}-im\d{o}-..." in the translation.\\
\textbf{Source:} However, <s>professional freelance photojournalists</s> are still quite rare. \\
\textbf{Target:} Sib\d{e}sib\d{e} <s>aw\d{o}n onise iroyin ti o ni im\d{o}-\d{e}r\d{o} ti o ni im\d{o}-\d{e}r\d{o} ti o ni im\d{o}-im\d{o}-im\d{o}-im\d{o}-im\d{o}-im\d{o}-iim\d{o}-im\d{o}-im\d{o}-im\d{o}-im\d{o}-im\d{o}-...</s> \\

Based on the 1 error detected, \textbf{the score of translation is: 0.0} \\

\bottomrule
\end{tabular}
}
\caption{Examples of Error-Span Predictions by different LLMs, ratings are from 0 to 1. The mistakes of the translation model is in \textcolor{red}{red}.}
\label{tab:qualitative}
\end{table*}

\begin{figure*}[htbp]
  \centering
  
  \makebox[\textwidth][c]{\includegraphics[width=1\textwidth]{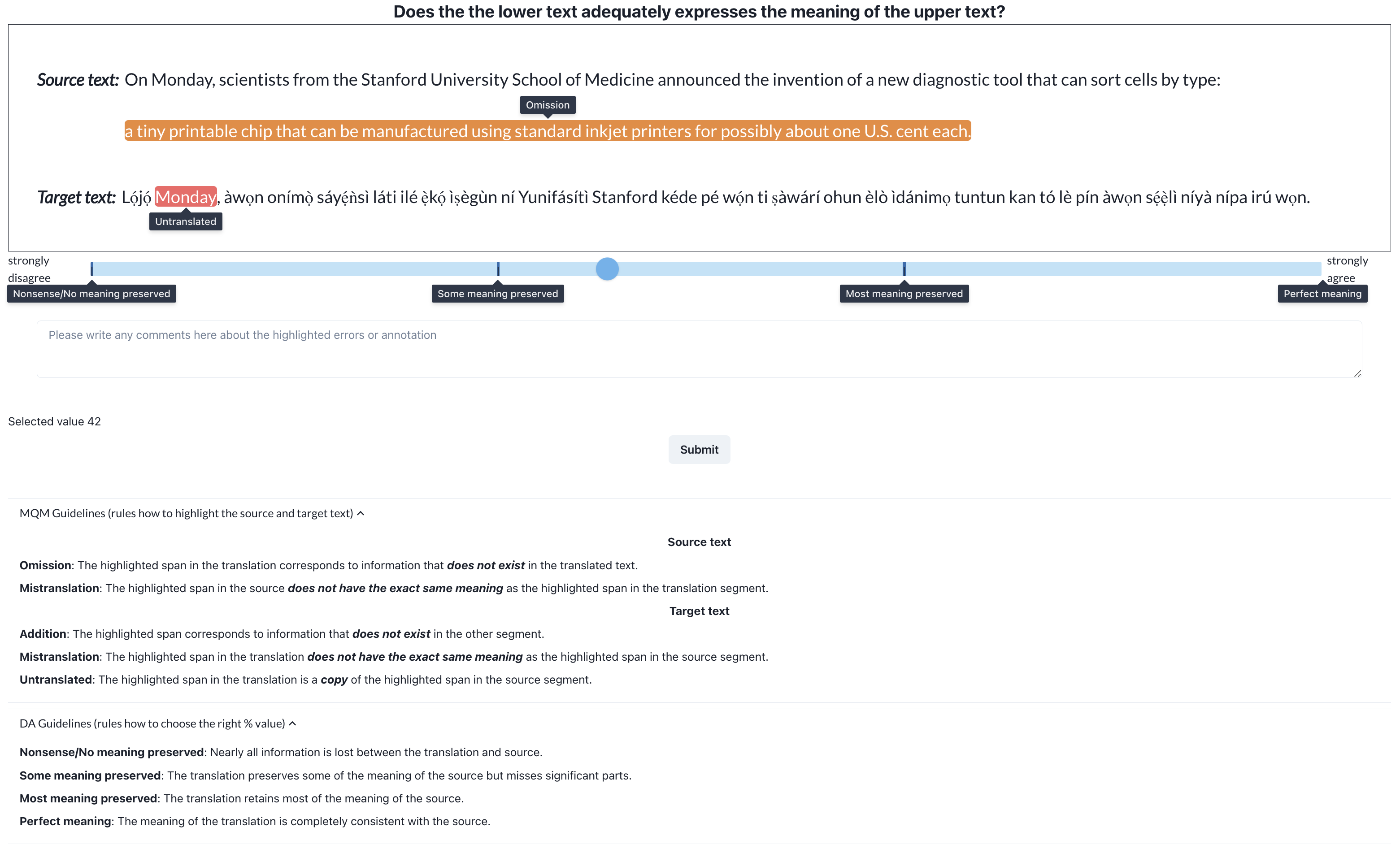}} 
  \caption{The annotation tool we used for the annotation process.}
  \label{fig:pdf-figure}
\end{figure*}
\label{appendix-annotation-tool}

\begin{figure*}[htbp]
  \centering
  
  \makebox[\textwidth][c]{\includegraphics[width=1\textwidth]{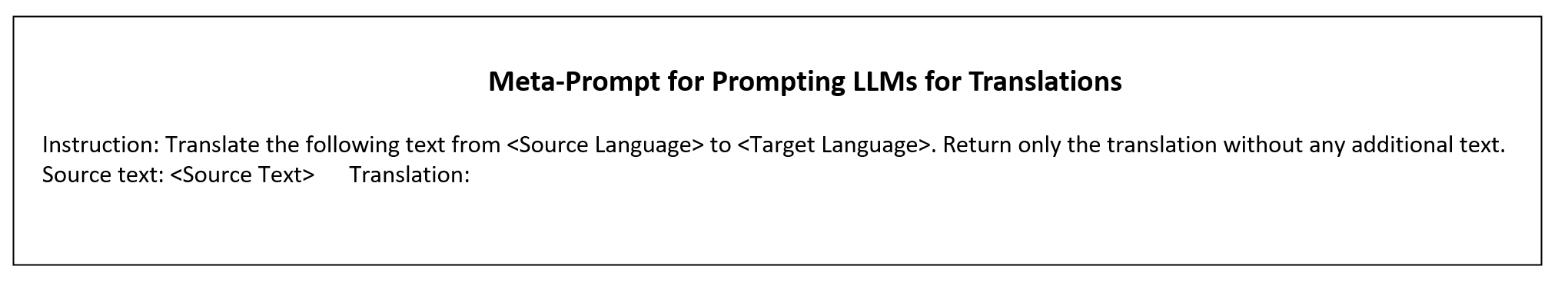}} 
  \caption{The prompt template used for prompting LLMs for translations.}
  \label{fig:prompt_translation}
\end{figure*}


\begin{figure*}[htbp]
  \centering
  
  \makebox[1\textwidth][c]{\includegraphics[width=1\textwidth]{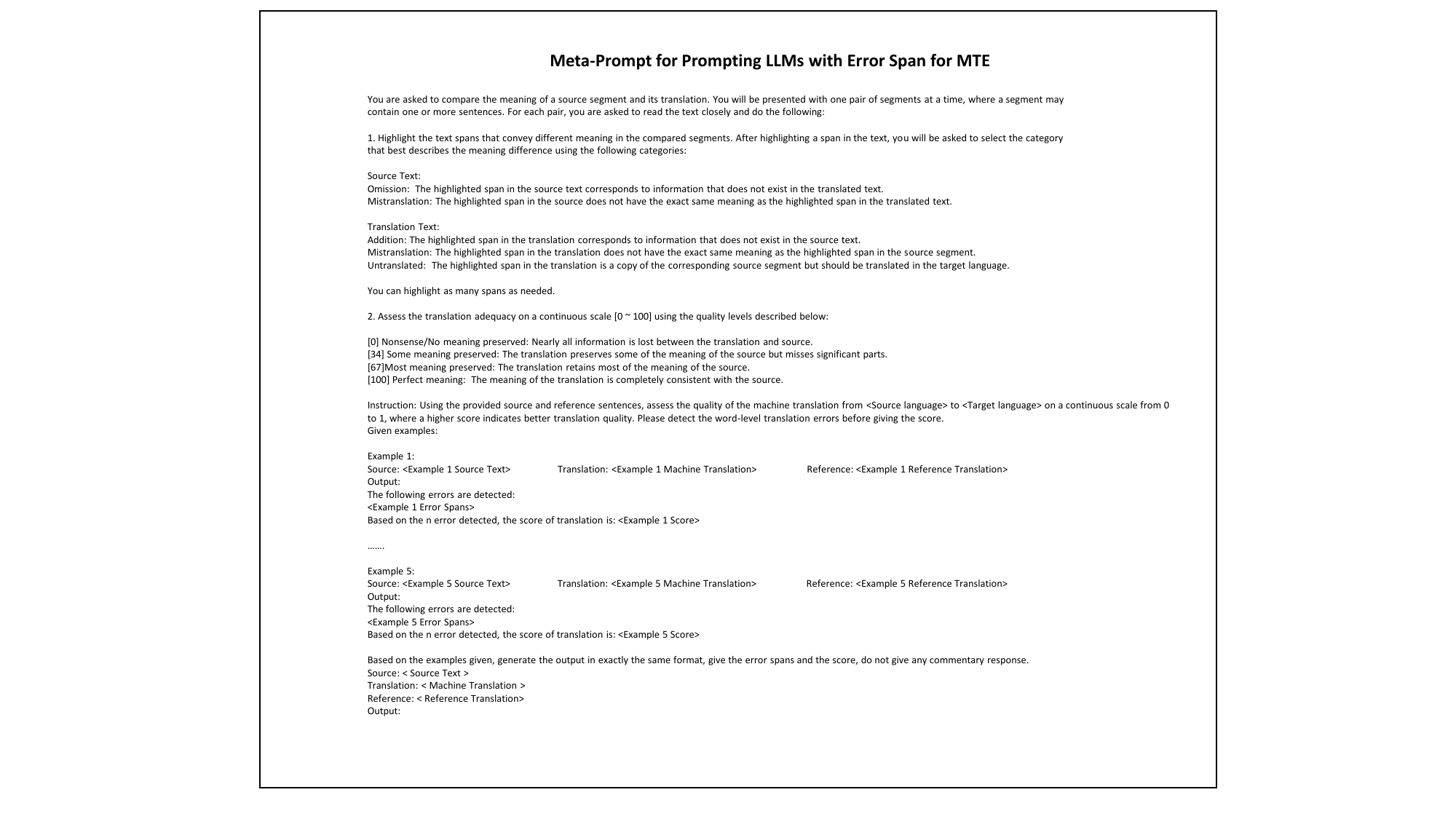}} 
  \caption{The prompt template used for prompting LLMs with error span detection for MTE.}
  \label{fig:template_with_error_span}
\end{figure*}


\begin{figure*}[htbp]
  \centering
  
  \makebox[\textwidth][c]{\includegraphics[width=1.1\textwidth]{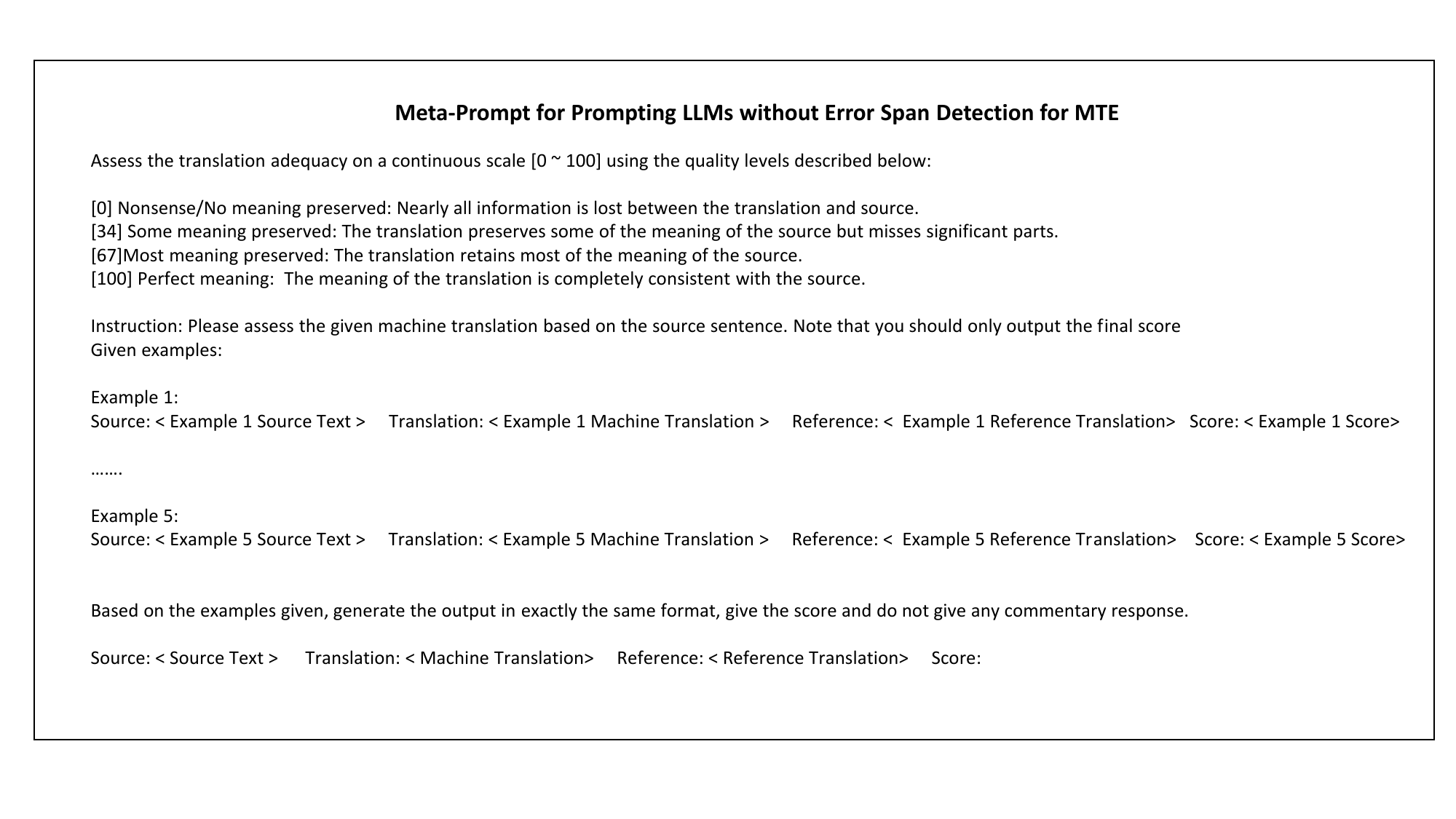}} 
  \caption{The prompt template used for prompting LLMs without error span detection for MTE.}
  \label{fig:llm-prompt-wo-MTE}
\end{figure*}

\begin{figure*}[htbp]
  \centering
  
  \makebox[\textwidth][c]{\includegraphics[width=1.1\textwidth]{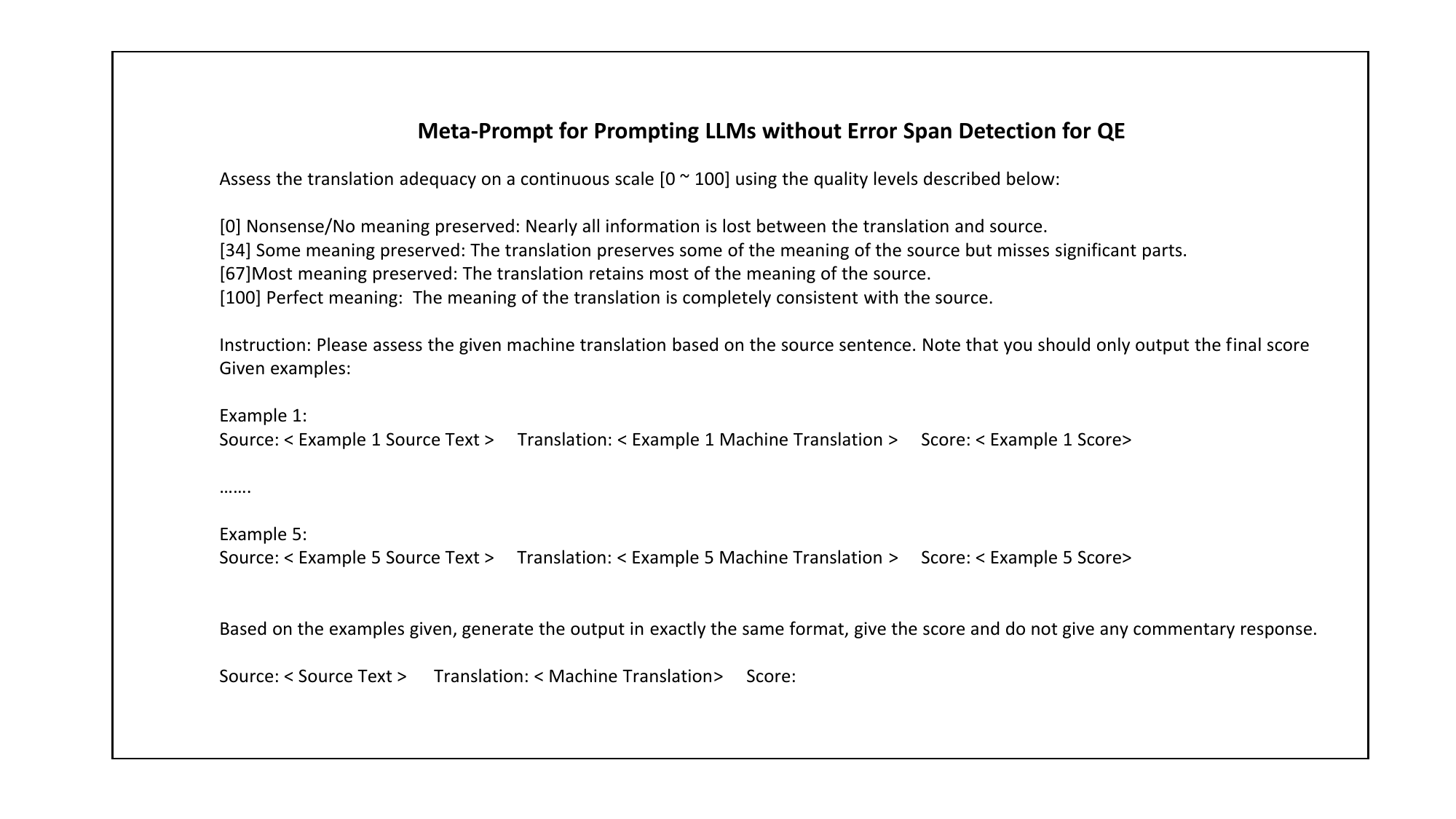}} 
  \caption{The prompt template used for prompting LLMs without error span detection for QE.}
  \label{fig:with_error_prompt_qe}
\end{figure*}



\label{appendix-annotation-guideline}

\begin{figure*}[htbp]
  \centering
  
  \makebox[\textwidth][c]{\includegraphics[width=1\textwidth]{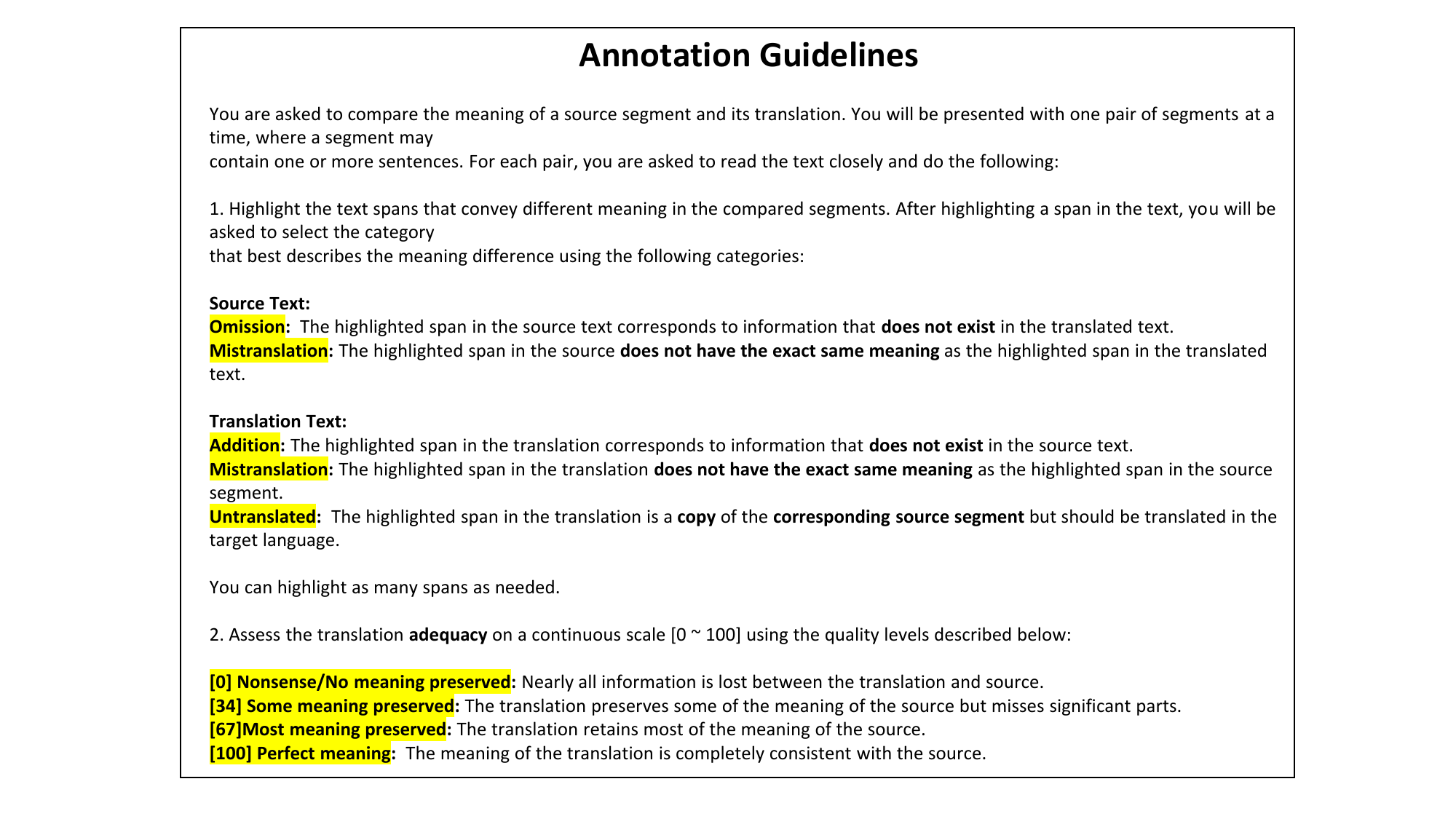}} 
  \caption{The annotation guideline we used for the annotation process.}
\end{figure*}


\begin{figure*}[htbp]
  \centering
  
  \makebox[\textwidth][c]{\includegraphics[width=0.6\textwidth]{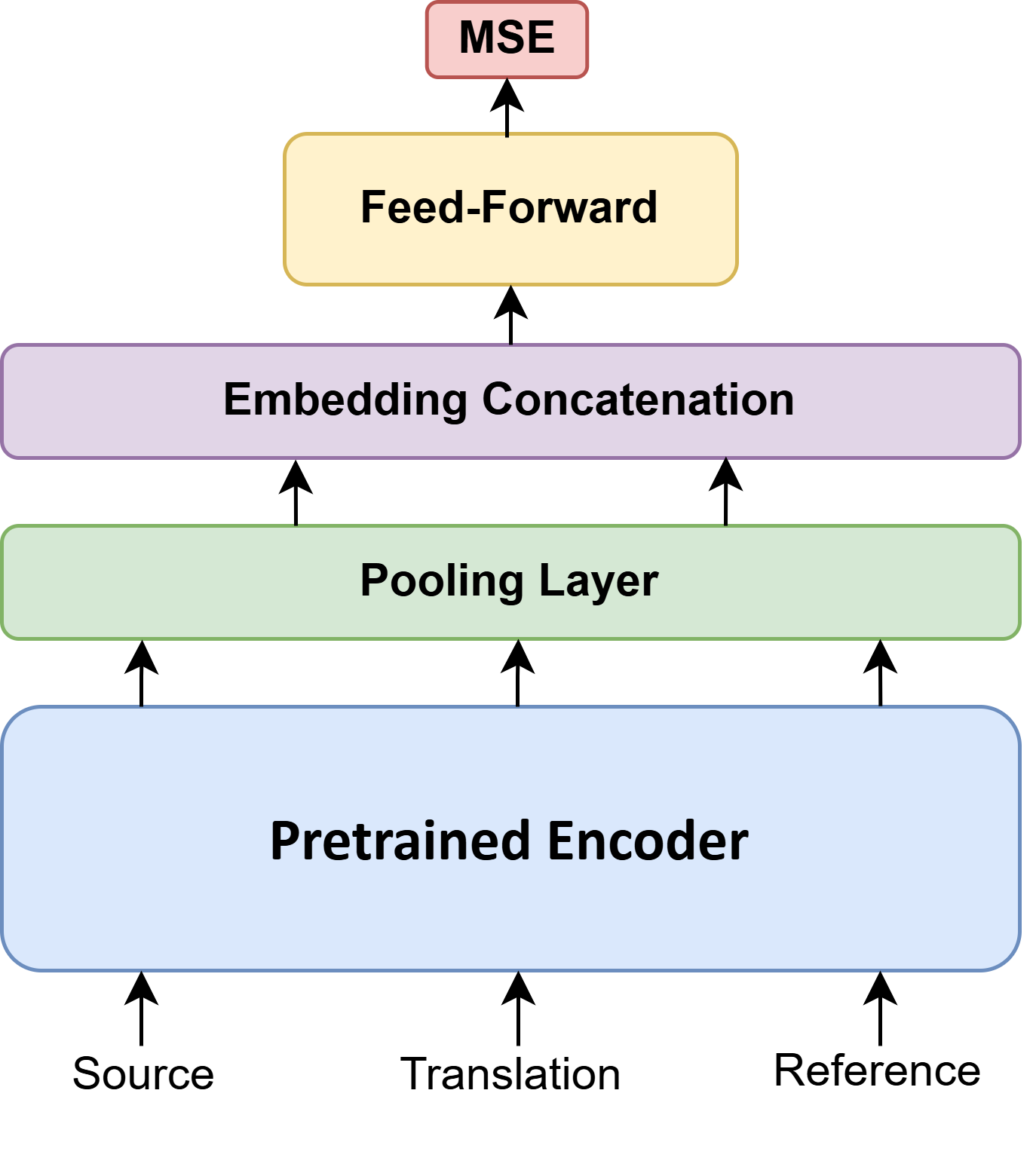}} 
  \caption{The workflow of the COMET architecture}
\end{figure*}

\end{document}